 \documentclass[final,5p,times,twocolumn]{elsarticle}

\usepackage{amssymb}
\usepackage{amsthm}
\usepackage{hyperref}
\usepackage{booktabs}
\usepackage{array} 
\usepackage{graphicx}
\usepackage{caption}
\usepackage{subcaption}
\usepackage{float}
\usepackage{tabularx}
\usepackage{array}
\usepackage{amsfonts}
\usepackage{geometry}
\usepackage{amsmath} 
\usepackage{makecell}
\usepackage{xcolor}

\journal{.}

\begin{document}

\begin{frontmatter}



\title{A longitudinal sentiment analysis of Sinophobia during COVID-19 using  large language models}


\author[inst2]{Chen Wang}  

\author[inst2]{Rohitash Chandra} 


\affiliation[inst2]{Translational Artificial Intelligence Research Group, School of Mathematics and Statistics, UNSW Sydney, Sydney, Australia}

\begin{abstract}
The COVID-19 pandemic has exacerbated xenophobia, particularly Sinophobia, leading to widespread discrimination against individuals of Chinese descent.  Large language models (LLMs) are pre-trained deep learning models used for natural language processing (NLP) tasks. %
The ability of LLMs to understand and generate human-like text make them particularly useful for analyzing social media data to detect and evaluate sentiments. 
We present a sentiment analysis framework utilising LLMs for longitudinal sentiment analysis of the Sinophobic sentiments expressed in X(Twitter) during the COVID-19 pandemic. 
The results show a significant correlation between the spikes in Sinophobic tweets, Sinophobic sentiments and surges in COVID-19 cases, revealing that the evolution of the pandemic influenced public sentiment and the prevalence of Sinophobic discourse. Furthermore, the sentiment analysis revealed a predominant presence of negative sentiments, such as "annoyed" and "denial" which underscores the impact of political narratives and misinformation shaping public opinion. 
The lack of empathetic sentiment which was present in previous studies related to COVID-19 highlights the way the political narratives in media viewed the pandemic and how it blamed the Chinese community. 
Our study highlights the importance of transparent communication in mitigating xenophobic sentiments during global crises.


\end{abstract}

\end{frontmatter}


\section{Introduction}
\label{sec:sample1}
 
The COVID-19 pandemic was a result of  severe acute respiratory syndrome coronavirus 2 (SARS-CoV-2) \cite{velavan2020covid,yuki2020covid} that broke out in December 2019 in Wuhan City, Hubei Province of China. Since then, COVID-19 spread rapidly throughout the world \cite{singhal2020review,ciotti2020covid} and to date (2024-05-21), there have been 704 million recorded cases and over 7 million deaths worldwide \cite{worldometer}. The pandemic had an effect on nearly all aspects of life, particularly in health, economy, and social interactions \cite{singhal2020review}.   Clemente-Suárez  et al. \cite{clemente2021impact} studied the psychological effects  created by COVID-19 and the economic downturn caused by the pandemic.  Long et al. \cite{long2022covid} examined the disruption in social relationships and health caused by COVID-19, noting both the positive aspects, such as the growth of community support in times of lockdown and negative aspects, including the rise of mental health issues, unemployment, and the drastic rise of economic and social inequality. Lopez-Leon et al. \cite{lopez2021more} presented a review of studies about COVID-19 and long-term effects, and reported that about 80\%    of infected COVID-19 patients developed one or more long-term symptoms. Furthermore, they identified long term impacts of COVID-19, with fatigue, headache, and attention disorder being the most common.

During the epidemic, social media played a vital role in the process of spreading information which had a significant impact on public behaviour, because of the high penetration rate, fast transmission spread, and wide coverage \cite{gao2020mental,tang2021important,hussain2020role,aggarwal2022role}. Well-known social media platforms such as   Twitter, Facebook, and Instagram featured information from official reports, misinformation, and rumours with conspiracy theories that promoted anti-vaccination and other related problems. Gao et al. \cite{gao2020mental} considered this information to be one of the causes of mental health issues and found that high exposure to social media was linked to increased anxiety and depression among users. Depoux et al. \cite{depoux2020pandemic} highlighted the need for effective communication strategies to manage public perception. The study also  highlighted that the spread of misinformation, most of the time accompanied by fearmongering and racism, was faster than the virus itself. Consequently, the Chinese diaspora  were one of the first targets of abuse and  discrimination \cite{gao2022sinophobia}.

Sinophobia, also known as anti-Chinese sentiment, refers to the fear and hatred of China and those of Chinese ancestry\cite{peyrouse2016discussing,bille2017sinophobia}. The COVID-19 pandemic has greatly exacerbated this form of prejudice—with an early appearance in the form of street graffiti and abusve in social media \cite{tahmasbi2021go}.  In later stages, Chinese restaurants were avoided, and a stop of housing Chinese students by homestay operators was also observed \cite{gao2022sinophobia}. Viladrich \cite{viladrich2021sinophobic} examined the stigmatisation of Chinese and East Asian populations during the COVID-19 pandemic and highlighted how the association of COVID-19 with China led to widespread racial stigma, fuelled by terms such as the "Chinese virus."

Large language models (LLMs) \cite{zhao2023survey,chang2024survey} consist of pre-trained deep learning models used for natural language processing (NLP) \cite{chowdhary2020natural,nadkarni2011natural} tasks. LLMs use deep learning models that are trained using a vast corpus of text data such as Wikipedia and possess the capacity to undertake a wide range of NLP tasks such as text summarizations, creative writing, and language translation \cite{kasneci2023chatgpt,chang2023survey}. LLMs can be applied to a range of domain applications, such as video understanding and interpretation \cite{zhang2023video}, medicine \cite{thirunavukarasu2023large,kitamura2023chatgpt}, education \cite{kasneci2023chatgpt}. Prominent LLM examples include  ChatGPT (Generative Pre-Trained Transformer) \cite{floridi2020gpt} and GPT-4 \cite{achiam2023gpt} by OpenAI,  BERT (Bidirectional Encoder Representations from Transformers) \cite{devlin2018bert} and Gemini \cite{team2023gemini} by Google. 

 Sentiment analysis, also known as opinion mining, is a technique used in  NLP to determine the emotional tone expressed by a body of text \cite{medhat2014sentiment,prabowo2009sentiment}. It is the process of detecting the emotions in the opinion or text and determines if the writer's attitude towards a specific topic is positive, negative or neutral \cite{yang2020senwave}. Sentiment analysis can be used to detect a wide range of emotions in text which is dependent on the training data.  Wankhade et al. \cite{wankhade2022survey} presented a survey that underlined the wide-ranging applications of sentiment analysis in fields such as marketing, finance, healthcare, and social media. The study  also pointed out the importance and versatility of sentiment analysis in today's data-driven world.
 Several studies focused on sentiment analysis to understand public emotions and reactions during this period. Chandra et al. \cite{chandra2021covid} presented a study on sentiment analysis of tweets from India during the rise of COVID-19. The study reported that the majority of tweets expressed optimistic sentiments, followed by annoyed and joking sentiments, and highlighted the effectiveness of LLMs in  predicting multiple sentiments in tweets. Lande et al. \cite{lande2023deep} explored the use of deep learning-based language models for topic modelling of COVID-19-related Twitter data in India, focusing on the Alpha, Delta, and Omicron variants. The methodologies used, such as data preprocessing, can inform a framework for our project. Furthermore,  Chandra et al. \cite{chandra2023analysis}   provided   analysis of antivaccine Twitter sentiments  and performed a longitudinal analysis of sentiments in vaccine-related tweets, comparing the trends with the number of COVID-19 cases and vaccination rates in selected countries.  
 
Despite the substantial body of research on Sinophobia during the COVID-19 pandemic, significant gaps remain. Existing studies tend to be theoretical or rely on surveys and other qualitative research methods based on a limited sample size. Such surveys often suffer from biases due to their reliance on self-reported data \cite{krumpal2013determinants}. Participants may not always be completely honest or objective in their responses, and their answers can be influenced by social desirability bias, recall bias, or the phrasing of questions. Moreover, while some research has utilised temporal analysis and word embeddings to analyze online Sinophobia, such as the study by Tahmasbi et al. \cite{tahmasbi2021go}, these analyses were confined to the early stages of the pandemic. Additionally, the sentiment analysis performed during COVID-19 has not specifically focused on Sinophobia. Studies by Chandra et al. \cite{chandra2021covid} and Lande et al. \cite{lande2023deep} have used deep learning models to analyze sentiments related to COVID-19 but primarily concentrated on data from India.  Chandra et al. \cite{chandra2023analysis} also studied vaccine-related sentiments and compared several countries in Asia-Pacific and highlighted the sentiments and phrases associated with vaccine hesitancy. Given the complex nature of Sinophobia  \cite{gao2022sinophobia} and the limitations presented, there is a significant gap in the comprehensive understanding of Sinophobic sentiments during COVID-19 over a broader timeline and across diverse geographic regions.

In this study, we investigate the dynamics of Sinophobia during the COVID-19 pandemic through sentiment analysis using selected LLMs. Our major goals include analysing the evolution of Sinophobic sentiments over an extended timeline during COVID-19  across selected countries in the Asia-Pacific, South America and Europe region. We fine-tune the  BERT model using a selected dataset (Senwave COVID-19 sentiment dataset \cite{yang2020senwave}) to detect a wide range of sentiments rather than only obtaining sentiment polarities. We aim to classify sentiments expressed in tweets related to Sinophobia, focusing on keywords such as "China virus" and "Wuhan virus," and calculate the polarity scores will be to determine the message's nature. This analysis will provide valuable insights and contribute to a better understanding of how such sentiments evolved in response to different phases of COVID-19, which include different types of lockdowns,  and phases of development and deployment of vaccines globally.  

We present the rest of the study  as follows: Section 2 provides an overview of some related work on sentiment analysis, Sinophobia and longitudinal analysis during the COVID-19 pandemic. Section 3 presents the methodology that includes  data, models and the implementation of sentiment analysis using LLMs. Section 4 presents the results  and  Section 5 discusses the implications of the findings. Section 6  concludes the paper with  key insights and directions for future research.

\section{Related work}

\subsection{LLMs for sentiment analysis}

  LLMs face challenges and limitations despite having successful applications \cite{samuelson2023generative,kaddour2023challenges}. Hadi et al. \cite{hadi2023large} conducted a comprehensive survey of LLMs and listed challenges such as model biases inherited from training data, high computational resource requirements, privacy concerns, and data security issues. These limitations are significant as they can affect the reliability of LLMs in different applications, including sentiment analysis. For instance, LLMs for sentiment analysis would give biased sentiment classifications if the training dataset contains biased language. This is a crucial issue since it may fuel stereotypes and misinformation, thus affecting the fairness of the sentiment analysis outcome.

  Hussein \cite{hussein2018survey} categorised challenges as theoretical such as handling of negation, spam and fake detection, and sarcasm detection.  In addition, the technical category considered effective feature extraction, multilingual processing and managing bipolar words, i.e. words that can have different meanings depending on context. Studies have also highlighted  the challenges in sentiment analysis. Wankhade et al. \cite{wankhade2022survey}  highlighted the difficulties of handling unstructured sentiments, which are found quite commonly in social media,  where the writers are not constrained by any regulations or guidelines. This involves dealing with slang, abbreviations, and emojis, which can significantly affect the accuracy of sentiment analysis models. These additions reflect how the field of sentiment analysis is evolving, particularly with the increasing prevalence of social media data.

In the realm of sentiment analysis, LLMs have demonstrated substantial advancements. Zhan et al. \cite{zhan2024optimization} noted the limitations of traditional sentiment analysis methods and emphasized the efficacy of GPT-3 in capturing complex emotional nuances through fine-tuning techniques, also achieving notable improvements in accuracy and contextual understanding. Deng et al. \cite{deng2023llms} utilised LLMs such as  GPT-3 and PaLM for market sentiment analysis on Reddit, demonstrating the effectiveness of in-context learning with Chain-of-Thought reasoning to generate sentiment labels for social media content. This study has underscored the transformative potential of LLMs in enhancing sentiment analysis.

 \subsection{Sinophobia}

Extensive research has documented Sinophobia during the COVID-19 pandemic \cite{viladrich2021sinophobic}. Chen et al. \cite{chen2020anti} presented a review that examines the rise in anti-Asian sentiment in the United States amid the COVID-19 pandemic, showing similarities to historical instances of racism against Asians. The study highlighted the society and healthcare contributions made by Asian Americans, while also addressing the recent increase in discrimination and hate crimes targeting Asians due to the pandemic. Gao \cite{gao2022sinophobia} provided a comprehensive framework using sociological, discursive and interpretive approaches for understanding the rise of Sinophobia during the COVID-19 pandemic. They reported that Sinophobia is driven not only by the stereotypical association of Chinese people with the coronavirus, but a complex phenomenon influenced by health, racial, and political factors. Masters-Waage et al. \cite{masters2020covid} conducted two studies at different time points during COVID-19, each with around 500 participants from the United States, Canada and India. These studies used Bayesian analysis and found no evidence that using place-specific names such as "Wuhan Virus" or "China Virus" increased Sinophobia. Cheah et al.\cite{cheah2020covid} focused on Chinese American families and collected data by self-reported surveys from hundreds of participants, and provided empirical evidence of the mental health impacts of COVID-19-related racial discrimination. Tahmasbi et al. \cite{tahmasbi2021go} analysed two large datasets from Twitter and \textit{4chan's}\footnote{4chan is anonymous English-language image board. Website: \url{https://www.4chan.org/}} Politically Incorrect board (pol) over a five-month period by using temporal analysis and word embedding to examine the dissemination and evolution of Sinophobic content. This study also identified several Sinophobic slurs that were used frequently on Twitter, which can serve as filtration terms in our study. 

\subsection{Longitudinal Analysis}

Longitudinal analysis is a crucial research method that focuses on studying changes over time by repeatedly observing the same subjects \cite{diggle2002analysis} allowing researchers to track changes and identify trends. Longitudinal analysis is widely used across different fields, including psychology \cite{liu2016longitudinal}, medicine \cite{neufeld1981clinical}, and social sciences \cite{newsom2013longitudinal}.
Related to our study, longitudinal analysis by Lucas et al. \cite{lucas2020longitudinal} on the immunological response in severe COVID-19 cases helped in understanding complex temporal dynamics that identified key patterns and correlations between immune response profiles and disease severity. Wang et al. \cite{wang2020longitudinal} conducted a study utilising longitudinal analysis for understanding the evolving psychological impact of the COVID-19 pandemic on a selected population in China. They highlighted how continuous monitoring can reveal persistent and emerging mental health issues despite interventions.


Chandra and Krishna \cite{chandra2021covid}   provided longitudinal sentiment analysis of COVID-19 tweets in India, highlighting the fluctuations in public mood corresponding to the rise and fall of new cases.   Chandra et al. \cite{chandra2023analysis} also employed longitudinal analysis to examine 
sentiments towards COVID-19 vaccines, covering   the early stages of the pandemic to the widespread rollout of vaccination programs. This research demonstrated how sentiment polarity fluctuated with key events, such as the announcement of vaccine efficacy results and the emergence of new variants.

\section{Methodology}

\subsection{Data}

\subsubsection{Global COVID-19 X (Twitter) dataset}

 The  Global COVID-19 Twitter dataset provides an extensive collection of tweets related to the COVID-19 pandemic from March 2020-February 2022 \cite{chandra2024global} and is publicly available on Kaggle. This dataset includes a wide array of tweets from six different countries: Australia, Brazil, India, Indonesia, Japan and the United Kingdom (as shown in Table \ref{tab:count}). It encompasses a wide range of public reactions and sentiments regarding the pandemic, allowing for complete analysis across various cultural and regional contexts. This dataset has also been employed by Chandra et al. \cite{chandra2023analysis} to analyse vaccine-related sentiments, demonstrating its applicability in studying public sentiment across different contexts. By leveraging this dataset, we can compare our findings with those from previous studies, ensuring consistency and robustness in our sentiment analysis framework.

\begin{table}[htbp!]
\centering
\small 
\begin{tabular}{lc}
\toprule
\textbf{Country} & \textbf{Tweets Count} \\
\midrule
Australia        & 3212464 \\
Brazil           & 943913  \\
India            & 5411294 \\
Indonesia          & 229935  \\
Japan             & 644510  \\
United Kingdom    & 16958866  \\
\bottomrule
\end{tabular}
\caption{The tweet count for six different countries.}
\label{tab:count}
\end{table}
 The preprocessing of the datasets is crucial to ensure clean and standardised input for the BERT model. The preprocessing steps involve various tasks such as lowercasing text, expanding contractions, removing hashtags and mentions, and handling emojis and abbreviations. We ensured consistency in data preparation, and transformed emojis into text and other non-text symbols and data. Table \ref{table:standardized_language} shows how the abbreviations and emojis are preprocessed, and Table \ref{table:examples} demonstrates some examples of tweets before and after preprocessing.

\begin{table*}[ht]
\centering
\small
\begin{tabular}{|p{7cm}|p{7cm}|}
\hline
\textbf{Tweet language} & \textbf{Standardised word} \\ \hline
ain’t & am not \\
i’ll’ve & I will have \\

lol & laughing out loud \\
u2 & you too \\ 
rt & retweet \\ 
asap & as soon as possible \\ 
COVID-19 & coronavirus \\ 
\includegraphics[height=1em]{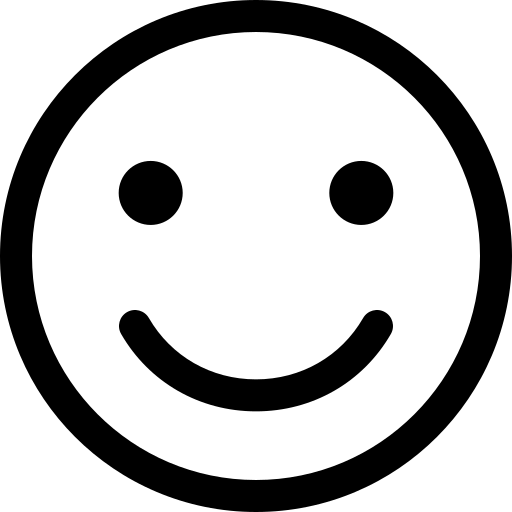} & smile \\ 
\includegraphics[height=1em]{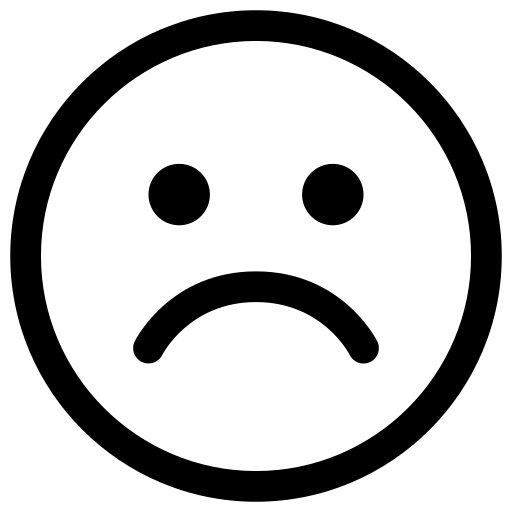} & sad \\ \hline
\end{tabular}
\caption{Changing tweet language usage and emojis to standardised language usages with semantic meaning.}
\label{table:standardized_language}
\end{table*}

\begin{table*}[ht]
\centering
\small
\begin{tabular}{|p{7cm}|p{7cm}|}
\hline
\textbf{Original Tweet} & \textbf{Preprocessed Tweet} \\ \hline

"Lol.....happy new month bro,quickly sending me something before coronavirus wipe April fool finally 0124152280 gtbank" & "laughing out loud \ \ \ \ \ happy new month bro quickly sending me something before coronavirus wipe april fool finally 0124152280 gtbank" \\ \hline

"@johnsy123aus If ScoMo is a Christian he wouldn't be looking to get involved in war games with China. We all  have enough to contend with in COVID times,without fighting." &  "if scomo is a christian he would not be looking to get involved in war games with china  we all  have enough to contend with in coronavirus times without fighting" \\ \hline

"Unlike other central banks which regulate policy through interest rates, the \#PBOC uses their \#RRR facility to maintain the monetary policy.

\#orbex\_fx \#investing \#trading \#forex \#forextrader \#forextrading \#China \#inflation\#covid \#omicronvariant \#housing   https://t.co/nB8ozIWWq9" & "unlike other central banks which regulate policy through interest rates  the  pboc uses their  rrr facility to maintain the monetary policy 

 orbex fx  investing  trading  forex  forextrader  forextrading  china  inflation  coronavirus  omicronvariant  housing" \\ \hline

"China is looking to create its first permanent military presence on the Atlantic Ocean, on the coast of the small African nation Equatorial Guinea, according to a report in The @WSJ based on classified U.S. intelligence. 

\#GetVaccinated \#WearAMask" & "china is looking to create its first permanent military presence on the atlantic ocean  on the coast of the small african nation equatorial guinea  according to a report in the  based on classified u s  intelligence  

 getvaccinated syringe" \\ \hline
\end{tabular}
\caption{Examples of tweets before and after preprocessing}
\label{table:examples}
\end{table*}

\subsubsection{SenWave dataset}

In this study, we trained and tested the BERT model by using the SenWave dataset \cite{yang2020senwave} and it is openly available on GitHub \cite{senwave2024}. The SenWave dataset compiles more than 104 million tweets and Weibo messages on COVID-19 in six Different languages: English, Spanish, French, Arabic, Italian, and Chinese. The data was collected between March 1 and May 15, 2020. It is specifically designed for analyzing social media posts related to COVID-19 and features 10,000 English and 10,000 Arabic tweets labeled into ten specific sentiment categories: optimistic, thankful, empathetic, pessimistic, anxious, sad, annoyed, denial, official report, and joking. This dataset has a wide coverage of public discourse during the pandemic and the labeling allows for a comprehensive analysis of the varied emotional responses, including the detection of Sinophobic sentiments. Furthermore, the effectiveness of the SenWave dataset has been demonstrated in two other studies mentioned in the introduction \cite{chandra2021covid,chandra2023analysis}. These studies also utilised this dataset and achieved promising results, underscoring the dataset's robustness and reliability for sentiment analysis tasks. After the BERT model is trained, we apply sentiment analysis on the Global COVID-19 X (Twitter) dataset.

\subsection{BERT-based model}

BERT (Bidirectional Encoder Representations from Transformers) is a state-of-the-art language representation model developed by Devlin et al. \cite{devlin2018bert}.  BERT has been pre-trained on a large corpus of text from the English Wikipedia and BookCorpus using two unsupervised tasks including Masked Language Modelling (MLM) and Next Sentence Prediction (NSP). Unlike traditional unidirectional language models, BERT adopts a deep bidirectional approach, allowing it to simultaneously consider both the left and right context of text. BERT employs the Transformer deep learning model  which utilises a self-attention mechanism to capture  nuanced relationships between words in a sentence. 

Fine-tuning BERT involves adapting the pre-trained model using task-specific data. BERT has been fine-tuned on various NLP tasks,  such as modelling US general elections \cite{chandra2021biden}, demonstrating its versatility and effectiveness \cite{koroteev2021bert,miranda2023exploring}. There are industry-specific implementations of BERT such as Legal-BERT \cite{chalkidis2020legal,shao2020bert} and Medical-BERT \cite{rasmy2021med,liu2021med}, using domain-specific data. There are several challenges in fine-tuning BERT models. The performance of fine-tuning is highly sensitive to hyperparameters such as learning rate, batch size, and number of epochs. Additionally, fine-tuning involves 
addressing challenges such as handling sarcasm, understanding nuanced sentiments, and ensuring the model generalises effectively from training data to unseen data.
In our study, by fine-tuning BERT on the SenWave dataset, which includes tweets labelled into ten emotions, we leverage its deep contextual understanding to accurately classify sentiments in social media posts.

\subsection{Framework}

\begin{figure*}[htbp]
    \centering
    \includegraphics[width=\textwidth]{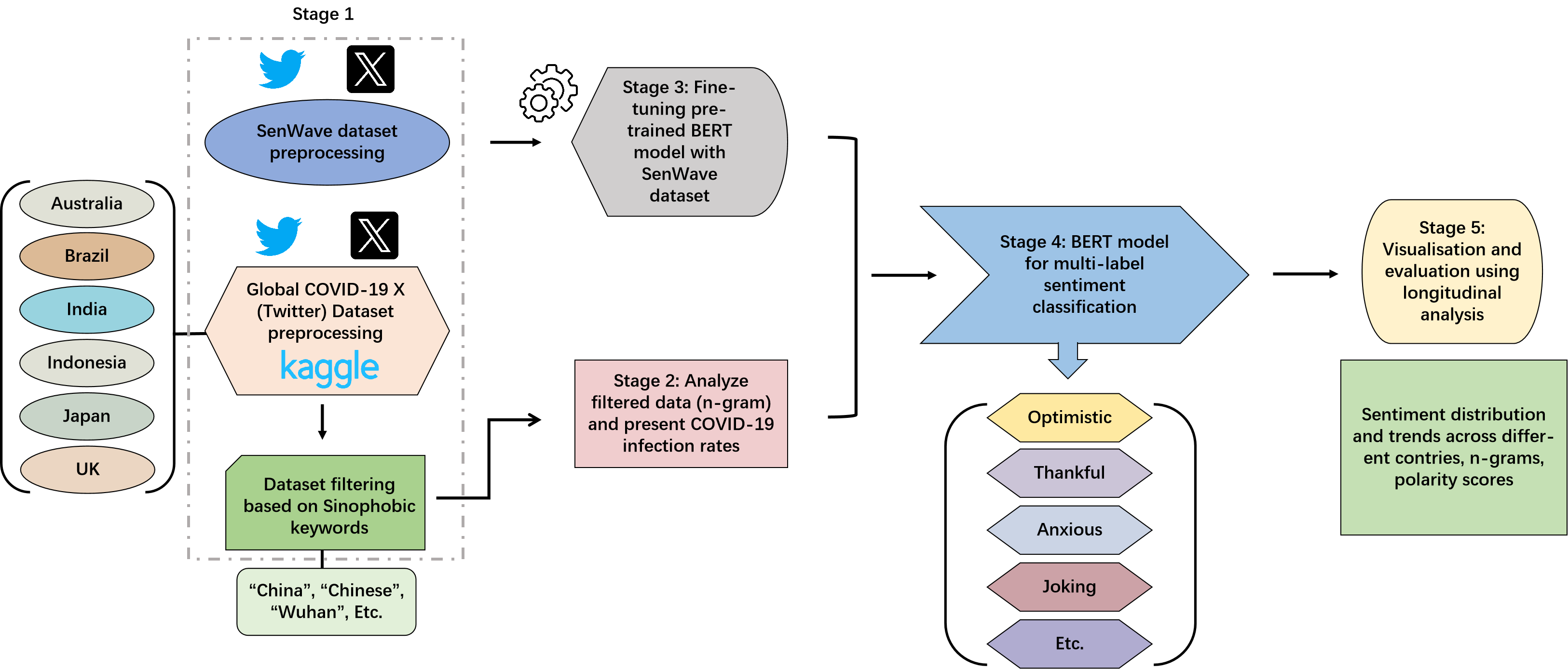}
    \caption{Framework for longitudinal sentiment analysis of Sinophobic tweets during COVID-19, involving five major stages: dataset preprocessing and filtering (Stage 1), n-gram analysis and infection rates (Stage 2), fine-tuning BERT model (Stage 3), multi-label sentiment classification (Stage 4), and longitudinal analysis and visualisation (Stage 5).}
    \label{fig:Framework}
\end{figure*}

Our sentiment analysis framework involves multiple stages, as illustrated in Figure \ref{fig:Framework}. 

In Stage 1, we clean and preprocess the SenWave dataset, which includes tweets labelled into ten distinct emotion categories. We clean tweets by removing the web links, hashtags and profile tags, and expanding contractions and abbreviations. This step ensures the data is clean and suitable for refining the BERT model. We obtained the Global COVID-19 X (Twitter) dataset, which includes a diverse set of tweets from different countries. We filter the data using specific Sinophobic keywords including  "China", "Chinese", "Wuhan" and "Sinophobia". We also included several Sinophobic slurs that occurred frequently on Twitter from the study by Tahmasbi et al. \cite{tahmasbi2021go}, such as "cn", "chink", and "chingchong". Table \ref{tab:sinophobia_keywords} shows all the keywords we used to determine Sinophobic tweets. We concatenated the filtered tweets into a single dataset ensuring no duplicates remain. We clean and preprocess the dataset using the same ways as the SenWave datasets, as demonstrated in Table \ref{table:examples}.

\begin{table}[htbp!]
    \centering
    \caption{Keywords Used to Determine Sinophobia}
    \begin{tabular}{|c|c|c|c|}
        \hline
        china & chinese & sinophobia & sinophobic \\
        prc & wuhan & hubei & beijing \\
        kung flu & chn & cn & ccp \\
        yellow peril & chink & chinks & chingchong \\
        ching chong & gook & chyna & mainland \\
        mainlander & bugland & chines & mainla \\
        chinazi & bugmen & chankoro & insectoid \\
        \hline
    \end{tabular}
    \label{tab:sinophobia_keywords}
\end{table}

In Stage 2, we provide an analysis of the data using n-gram (bigram and trigram) analysis and visualisation of Sinophobic tweets extracted from different countries. We also present COVID-19 infection rates for other countries that would be used for further analysis in the final stage. 

In Stage 3, we utilise a pre-trained BERT model available through Hugging Face \footnote{Hugging Face is a platform providing open-source tools \footnote{Website: \url{https://huggingface.co/}}}, which provides open-source NLP tools.  In our study, we fine-tune the pre-trained BERT model for sentiment analysis of COVID-19-related tweets. This step involves fine-tuning the pre-trained BERT model using the cleaned and preprocessed SenWave dataset. The fine-tuning process utilises the Adam optimiser \cite{kingma2014adam} which is known for its efficiency in handling large deep learning models. The SenWave dataset is divided into training and testing subsets, with a split ratio of 90\% for training, and 10\% for testing. Each subset undergoes a series of transformations to convert the tweets into a format suitable for the BERT model. This involves tokenization, where each tweet is broken down into tokens using the BERT tokenizer, and padding, where sequences are padded to ensure uniform length. The model architecture comprises three layers. The process starts with initialising pre-trained weights from  BERT-base-uncased \footnote{\url{https://huggingface.co/tftransformers/bert-base-uncased}} which provides a rich contextual representation of the input tweets. Next, we add a dropout layer to prevent overfitting and then a linear layer that maps the output to the ten sentiment categories. During training, we iterate over the training data to present it it to the model in batches and compare the output against the actual labels using the binary cross-entropy  loss function. We use the Adam optimiser to adjust the model's weights to minimise the loss. We conduct the training for four epochs monitor the model's performance at regular intervals and evaluate the model's performance on the test dataset. During the evaluation, we pass the test data through the model and calculate various metrics, such as Hamming Loss, Jaccard Score, Label Ranking Average Precision Score, and F1 scores (macro and micro). These metrics help me understand the model's accuracy and effectiveness in classifying sentiments.

 Stage  4 involves multi-label sentiment analysis, where the fine-tuned BERT model is used to analyze the sentiment of tweets on a broader level, taking into account the context and nuances of each tweet. In the final visualisation and evaluation (Stage 5), we perform longitudinal analysis and generate various plots to analyze the sentiment distribution and trends across different regions, including sentiment polarity scores. We obtain the polarity scores using the TextBlob library and also utilise custom weight ratios for different sentiment labels, as shown in Table\ref{tab:weight}.

\begin{table}[htbp!]
\centering
\small 

\begin{tabular}{lc}
\toprule
\textbf{Sentiments} & \textbf{Weight ratios} \\
\midrule
Optimistic        & 3 \\
Thankful           & 2  \\
Empathetic            & 0 \\
Pessimistic          & -3 \\
Anxious             & -2  \\
Sad    & -2  \\
Annoyed    & -1 \\
Denial    & -4 \\
Official report    & 0 \\
Joking          & 1 \\
\bottomrule
\end{tabular}
\caption{Weight ratios for different sentiments for polarity score calculation.}
\label{tab:weight}
\end{table}


\subsection{Experimental setting}

We present the details for the implementation of our sentiment analysis framework including model topology, hyperparameters, and other relevant settings used during experimentation. Our sentiment analysis framework is implemented using Python.  We use Python-based libraries including PyTorch for deep learning, Transformers from Hugging Face for implementing pre-trained BERT model, TorchText for data processing, Matplotlib and
Seaborn for visualisation and various utility libraries such as Pandas, NumPy, and Scikit-learn for data manipulation and evaluation. 

  We used the learning rate of 1e-05 in the Adam optimiser (taken from the literature) and  refined the BERT  model  for four epochs,  sufficient to converge on the training dataset while preventing overfitting as shown in earlier works \cite{chandra2023analysis,chandra2022semantic}.  We utilised model training and evaluation experiments using a computer equipped with an NVIDIA GPU (graphic processing unit)   with an AMD Ryzen 7 5800H processor and 16 gigabytes of memory to expedite the training process.

\section{Results}

\subsection{Data Analysis}

\begin{figure}[htbp!]
    \centering
    \includegraphics[width=0.49\textwidth]{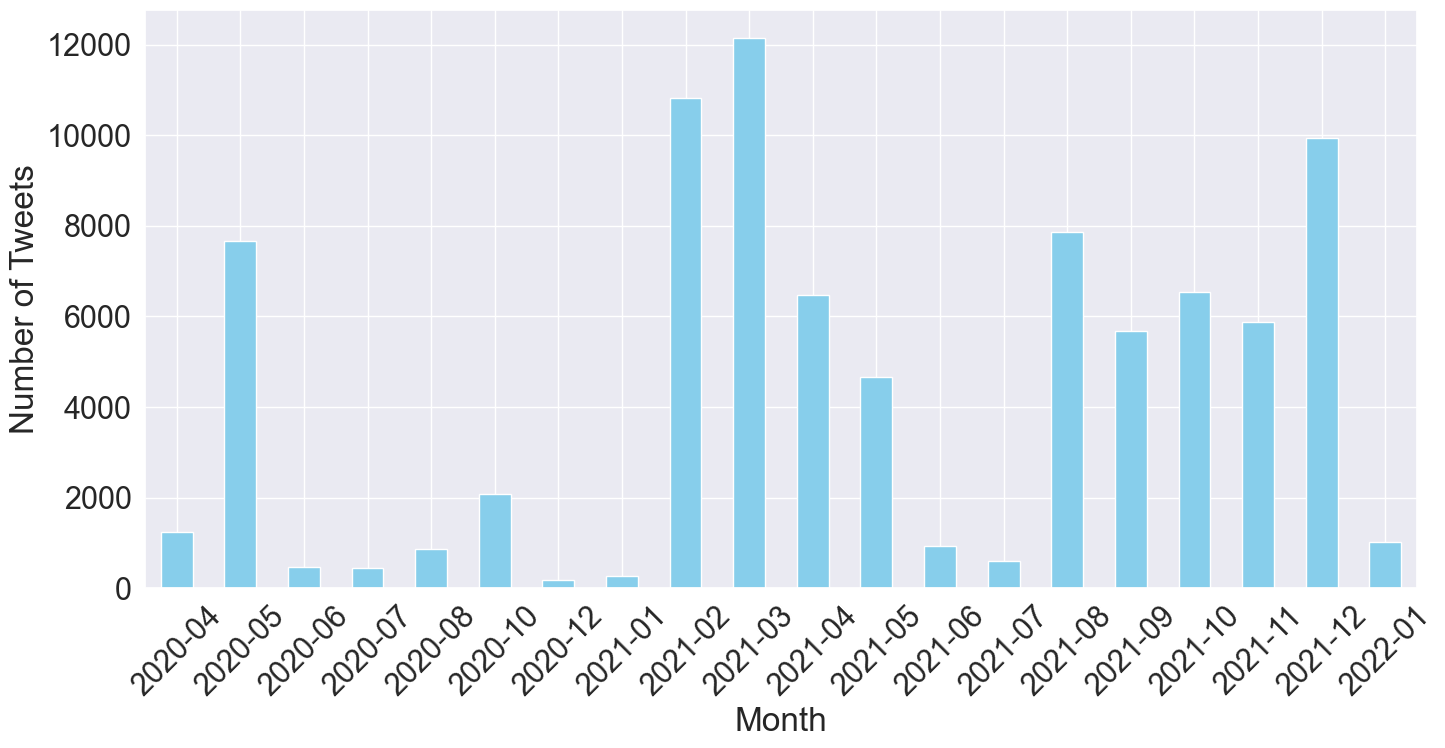}
    \caption{Number of total Sinophobic tweets over time}
    \label{fig:NumTweets}
\end{figure}

\begin{figure*}[htbp]
    \centering
    \includegraphics[width=\textwidth]{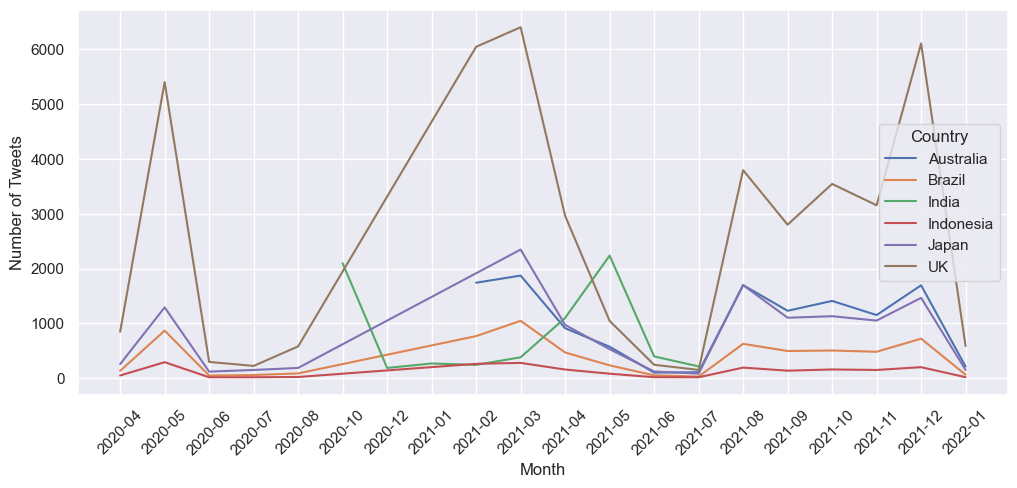}
    \caption{Number of Sinophobic tweets for each country over time.}
    \label{fig:NumTweetsOverMonth}
\end{figure*}

Firstly, we analyse the Sinophobic tweets extracted from the dataset as depicted in Figure \ref{fig:NumTweets} and Figure \ref{fig:NumTweetsOverMonth}, these tweets span from April 2020 to January 2022. The monthly distribution of these tweets reveals notable fluctuations in their frequency throughout the selected period.

\vspace{1cm}

\begin{figure*}[htbp]
    \centering
    \includegraphics[width=\textwidth]{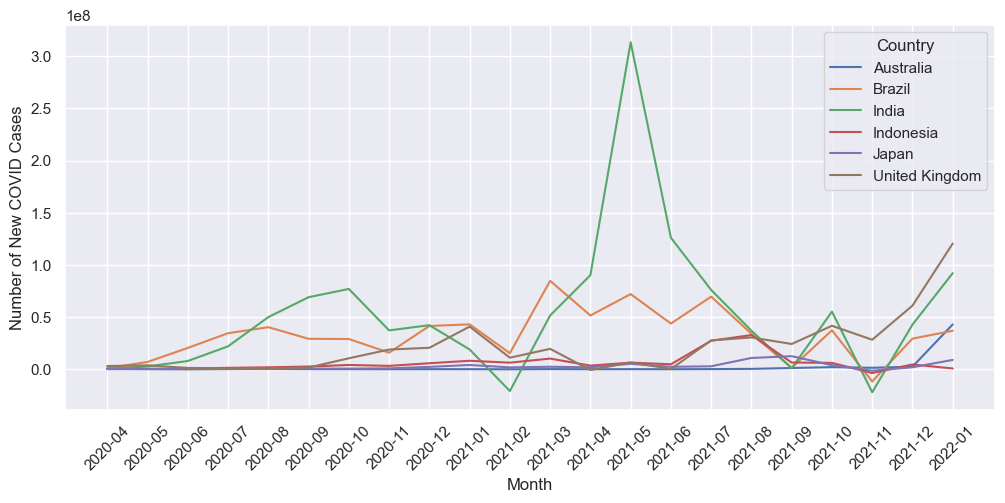}
    \caption{Number of COVID cases for each country over time.}
    \label{fig:COVIDcases}
\end{figure*}




Figure \ref{fig:NumTweets} shows a major spike in cases in May 2020, indicating a significant increase in Sinophobic tweets. Afterwards,  there is a noticeable drop in the number of Sinophobic tweets in the subsequent months, with relatively low activity during the summer of 2020. The overall distribution of tweets shows a gradual increase from the beginning of the pandemic, peaking in early 2021. This trend suggests that as the pandemic progressed with more information and discussion, there was a rise in Sinophobic sentiment on social media platforms. We further compare the monthly distribution of Sinophobic tweets with the number of new COVID-19 cases in various countries  (Figure \ref{fig:COVIDcases}), where the monthly new COVID-19 cases for Australia, Brazil, India, Indonesia, Japan, and the United Kingdom. The COVID-19 case data was sourced from Dong et al. \cite{dong2020interactive}, which is an online interactive dashboard that monitors reported cases of COVID-19 led by Johns Hopkins University.
There is a clear correlation between the spikes in Sinophobic tweets and the surges in COVID-19 cases. We find that the major peak in Sinophobic tweets in February and March 2021 aligns with a notable increase in COVID-19 cases, particularly in countries such as India. Similarly, the December 2021 peak coincides with another wave of COVID-19 cases, which could have influenced the rise in anti-Chinese sentiment.

\begin{figure}[htbp!]
    \centering
    \begin{subfigure}[b]{0.49\textwidth}
        \includegraphics[width=\textwidth]{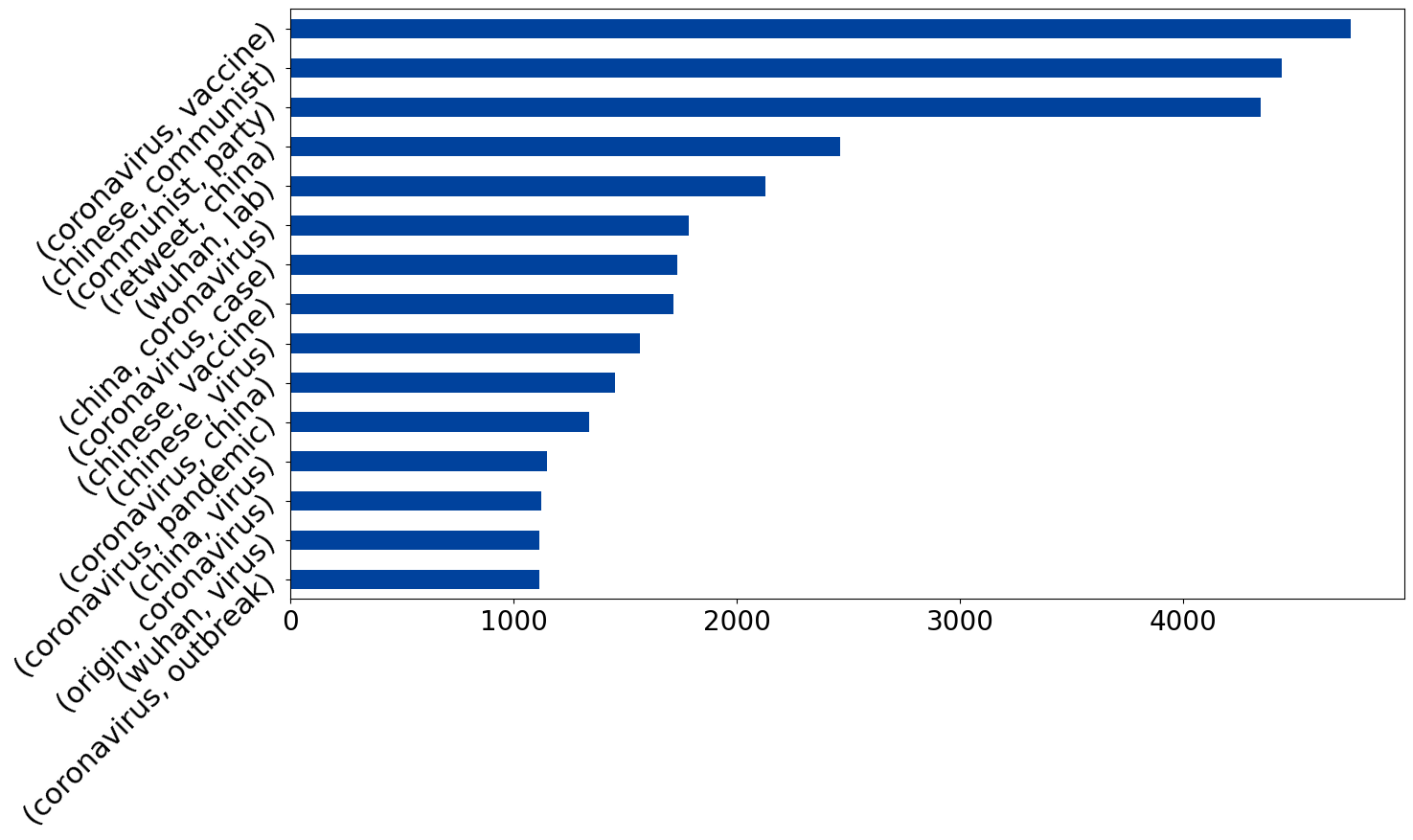}
        \caption{Bigram}
        \label{Bigram}
    \end{subfigure}
    \hfill
    \begin{subfigure}[b]{0.49\textwidth}
        \includegraphics[width=\textwidth]{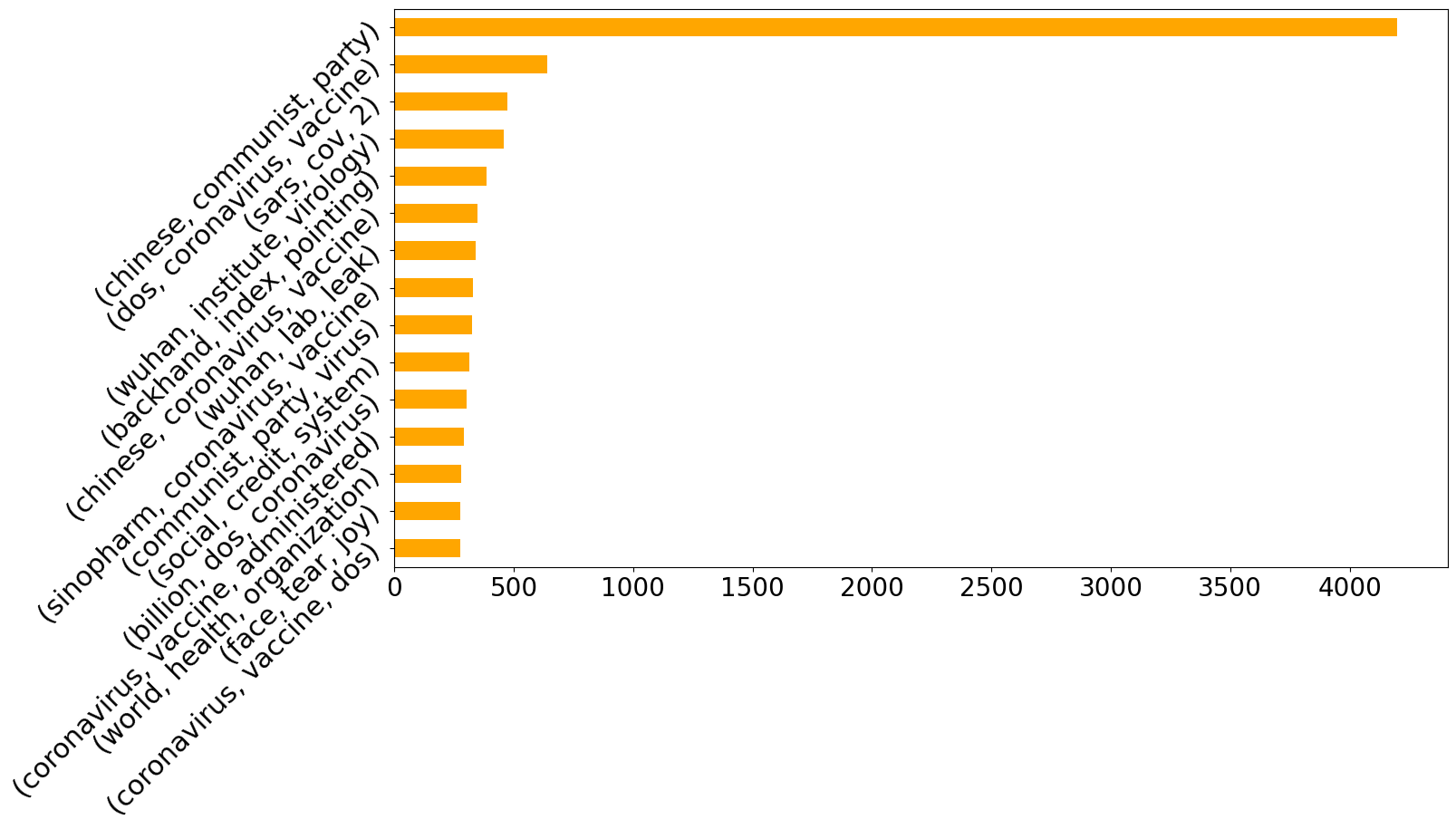}
        \caption{Trigram}
        \label{Trigram}
    \end{subfigure}
    \caption{Bigrams and trigrams of Sinophobic tweets between 2020-04 and 2022-01 worldwide.}
    \label{fig:bigram_trigram}
\end{figure}

We performed n-gram analyses \cite{brown1992class} to delve deeper into the content of the extracted Sinophobic tweets, focusing on bigrams and trigrams.  Figure \ref{fig:bigram_trigram} presents the top 15 bigram and trigram of the extracted tweets from April 2020 to January 2022 of all six countries.

As shown in Figure \ref{fig:bigram_trigram}-Panel (a), the most frequent bigrams in the dataset are "coronavirus vaccine", "chinese communist" and "communist party". The bigram "coronavirus vaccine" appears most frequently, indicating that discussions about the virus itself and the vaccine are central to the tweets. The bigrams "chinese communist" and "communist party" highlight a strong focus on the Chinese government and its ruling party. The  "wuhan lab", "chinese coronavirus" and "chinese vaccine" phrases indicate discussions related to the origin of the virus in Wuhan and the association of COVID-19 with China, contributing to the negative sentiments towards the country and its people.

In Figure \ref{fig:bigram_trigram}-Panel (b), the top China-related trigrams include "chinese communist party", "wuhan institute virology" and "chinese coronavirus vaccine". The trigram "wuhan institute virology" underscores the focus on the Wuhan Institute of Virology. This institute was at the centre of the COVID-19 lab leak theory, which is a controversial hypothesis suggesting that the virus accidentally escaped from the lab in Wuhan, China \cite{bbc2021labLeakTheory}. The terms "chinese communist party" and "chinese coronavirus vaccine" show tweets explicitly linked COVID-19 to China, thereby fuelling Sinophobic sentiments.

\begin{figure*}[htbp] 
\begin{subfigure}{0.48\textwidth}
\includegraphics[width=\linewidth]{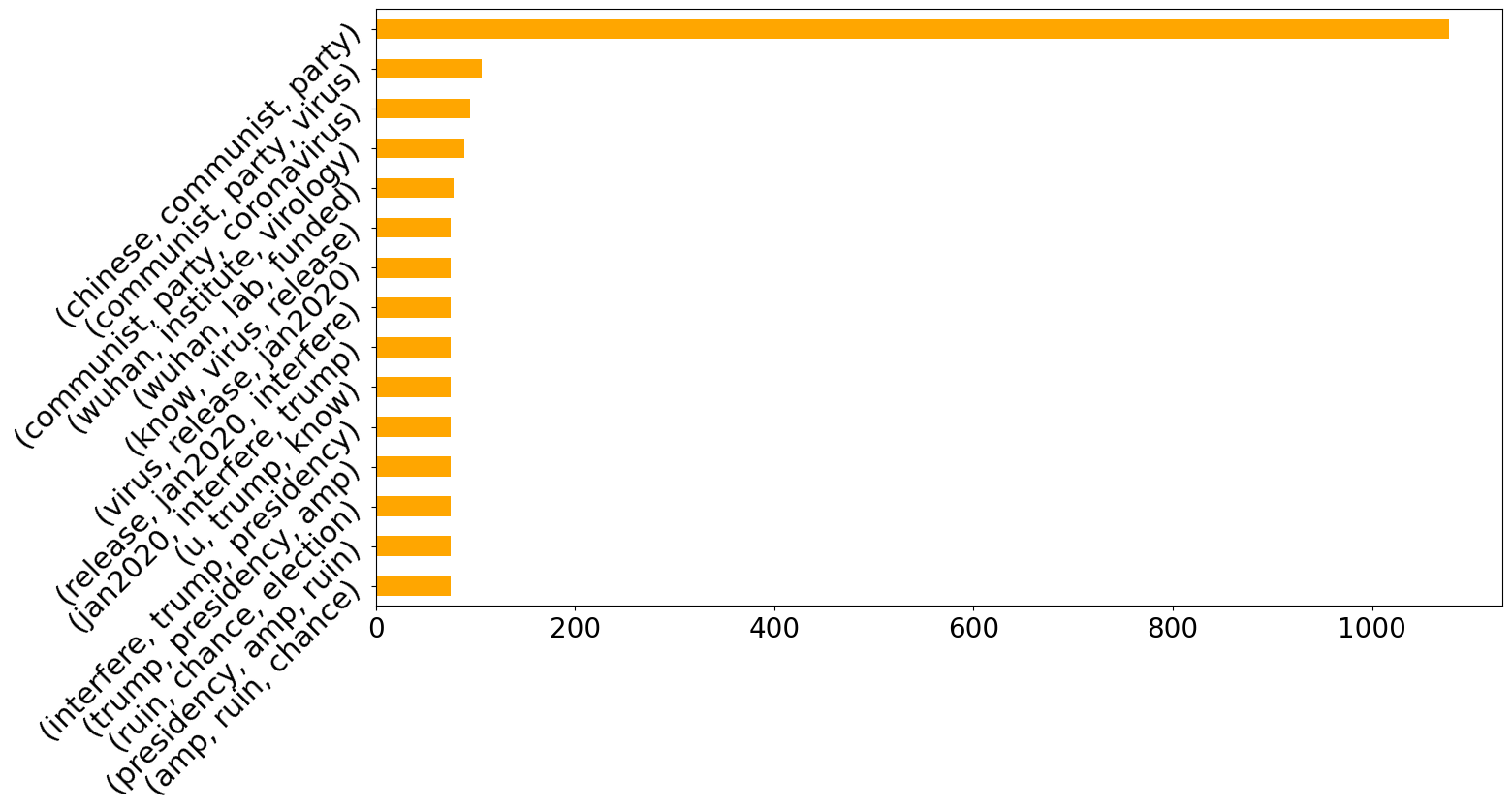}
\caption{Australia} \label{fig:a}
\end{subfigure}\hspace*{\fill}
\begin{subfigure}{0.48\textwidth}
\includegraphics[width=\linewidth]{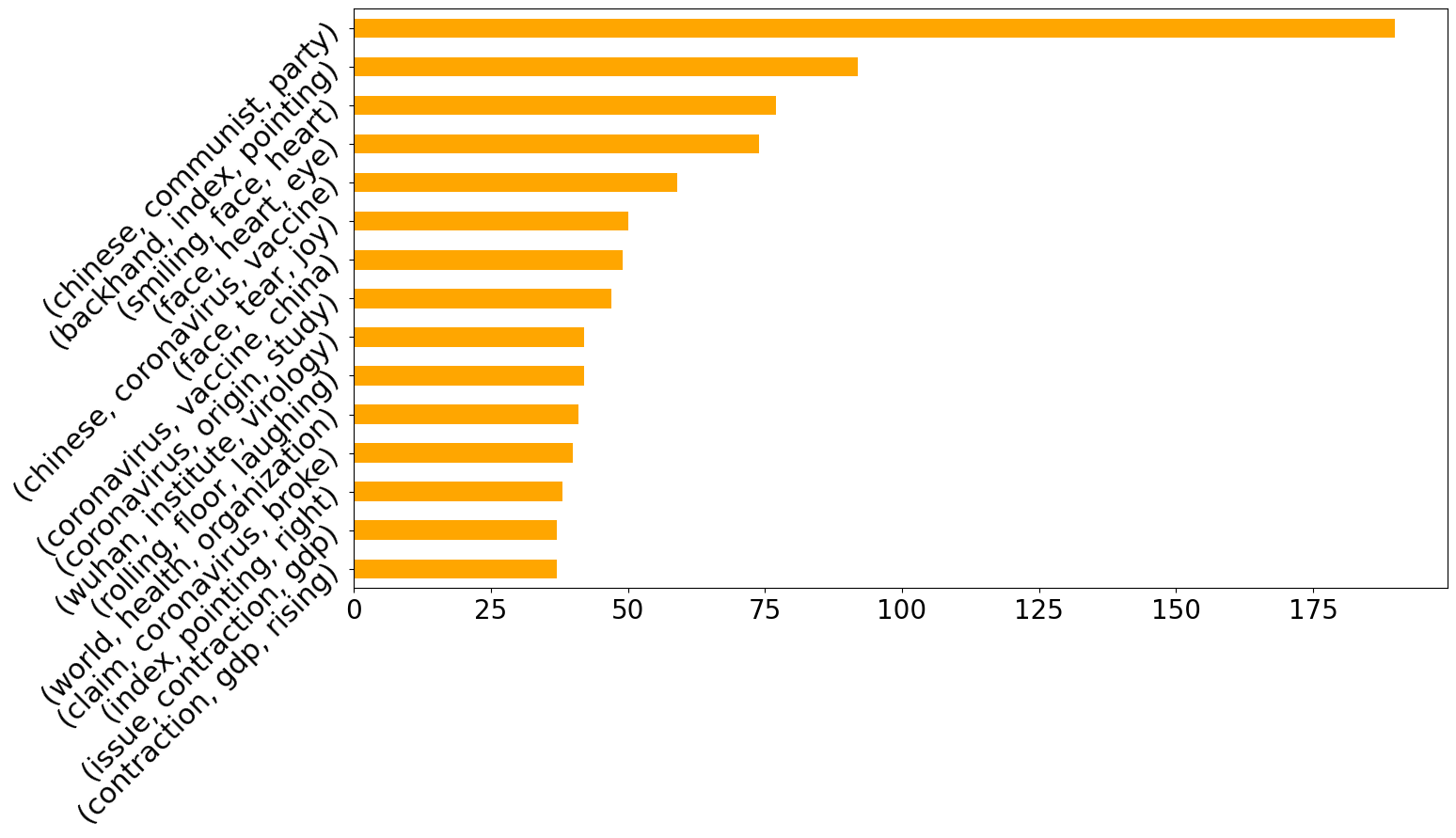}
\caption{India} \label{fig:b}
\end{subfigure}

\medskip
\begin{subfigure}{0.48\textwidth}
\includegraphics[width=\linewidth]{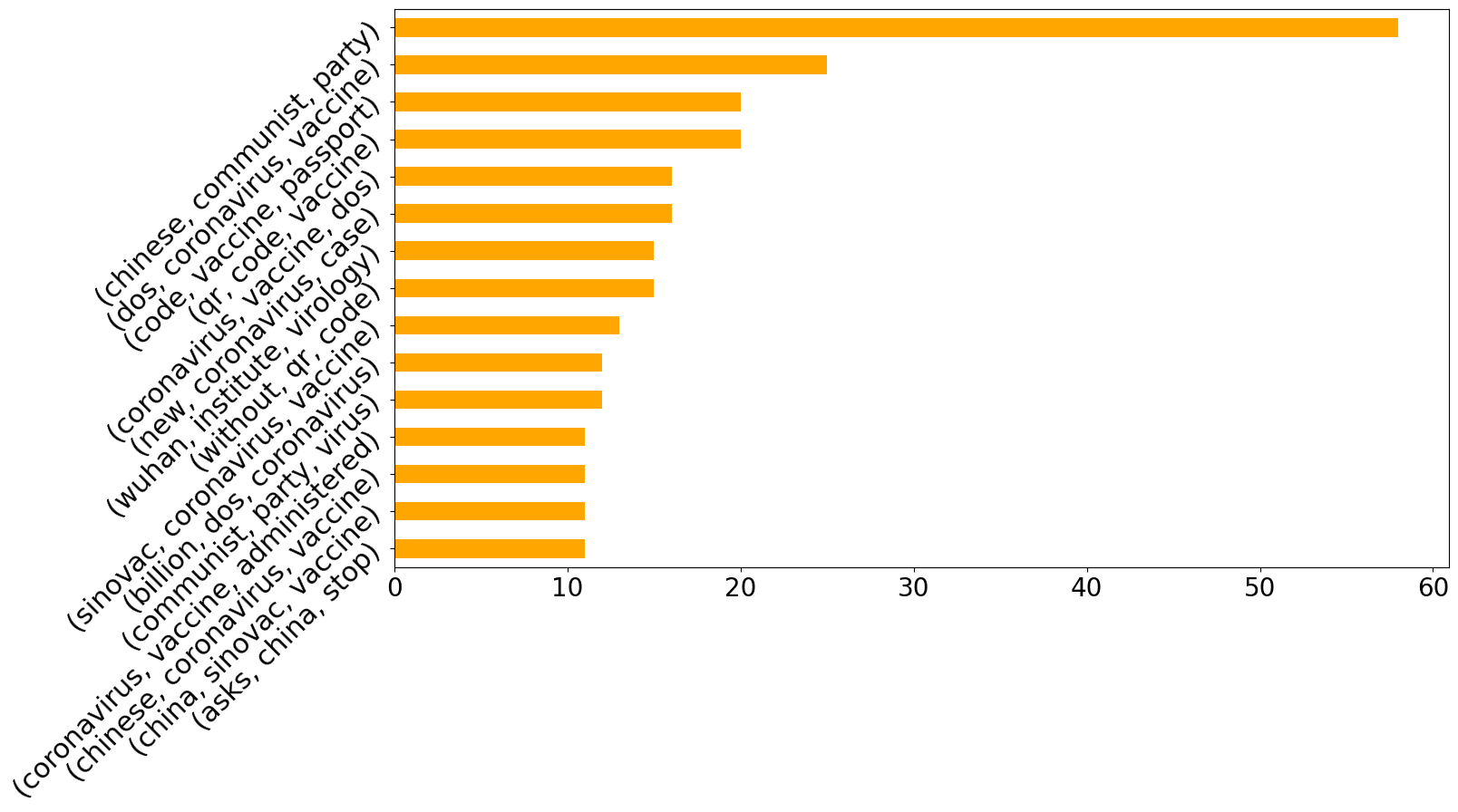}
\caption{Indonesia} \label{fig:c}
\end{subfigure}\hspace*{\fill}
\begin{subfigure}{0.48\textwidth}
\includegraphics[width=\linewidth]{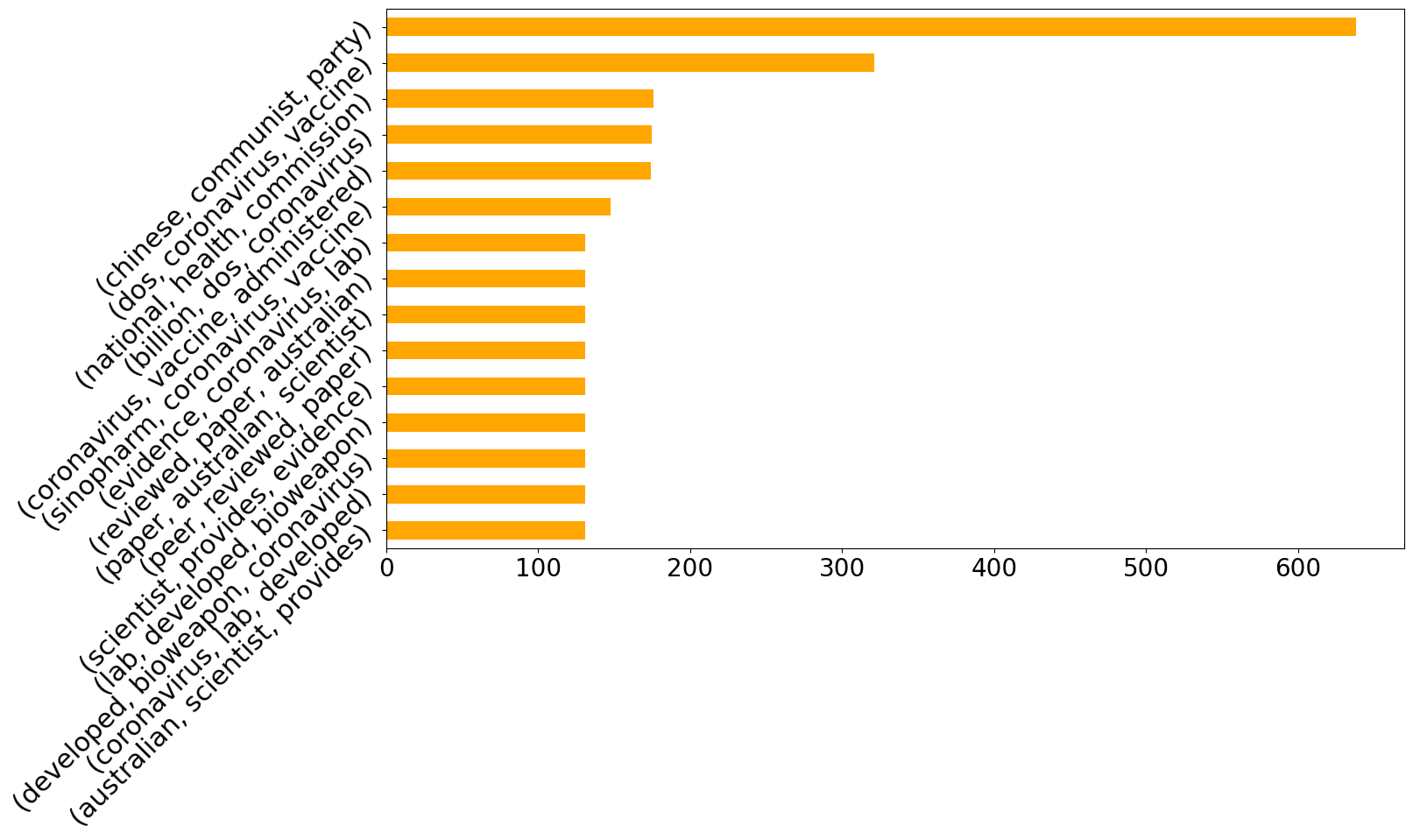}
\caption{Japan} \label{fig:d}
\end{subfigure}

\medskip
\begin{subfigure}{0.48\textwidth}
\includegraphics[width=\linewidth]{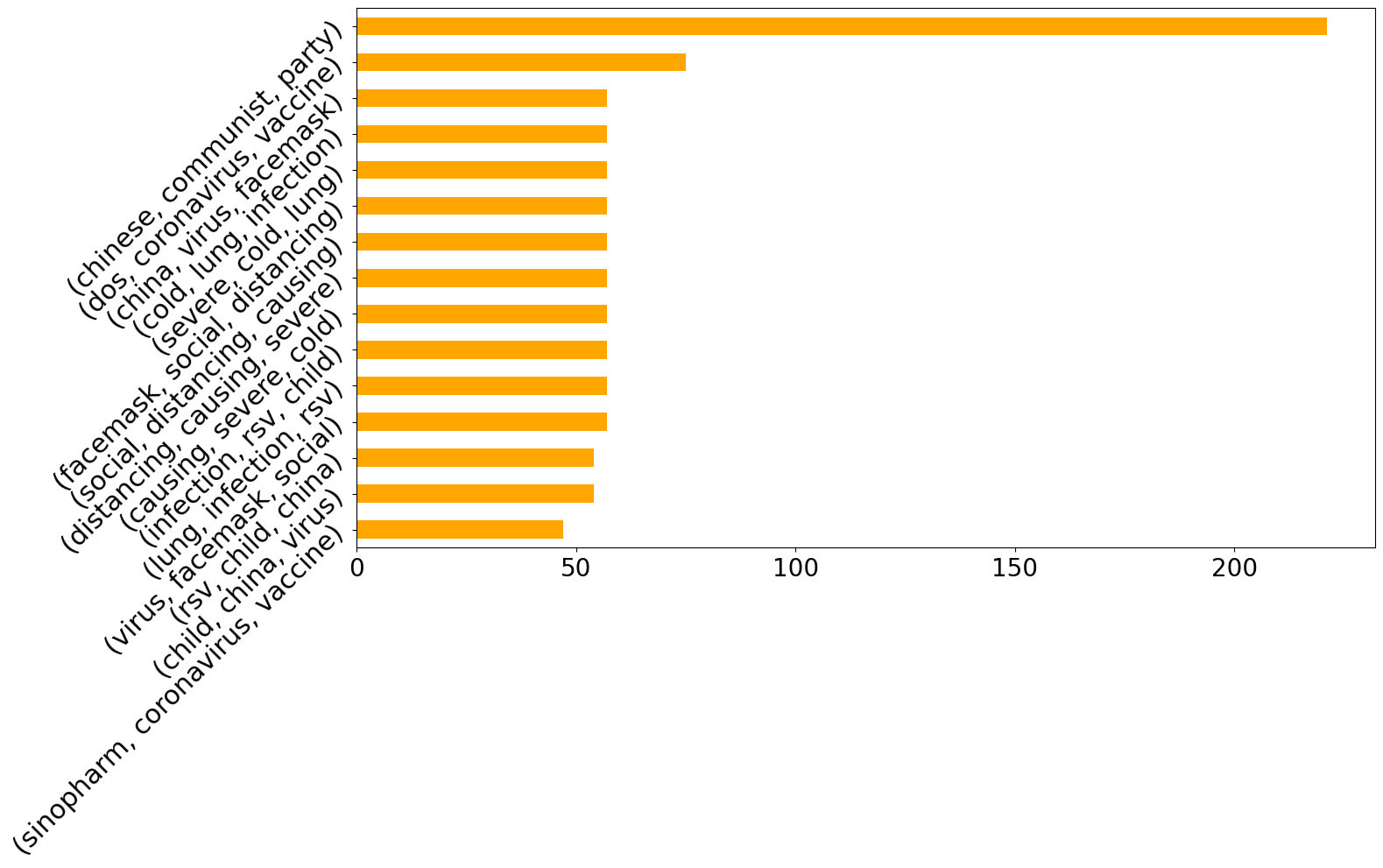}
\caption{Brazil} \label{fig:e}
\end{subfigure}\hspace*{\fill}
\begin{subfigure}{0.48\textwidth}
\includegraphics[width=\linewidth]{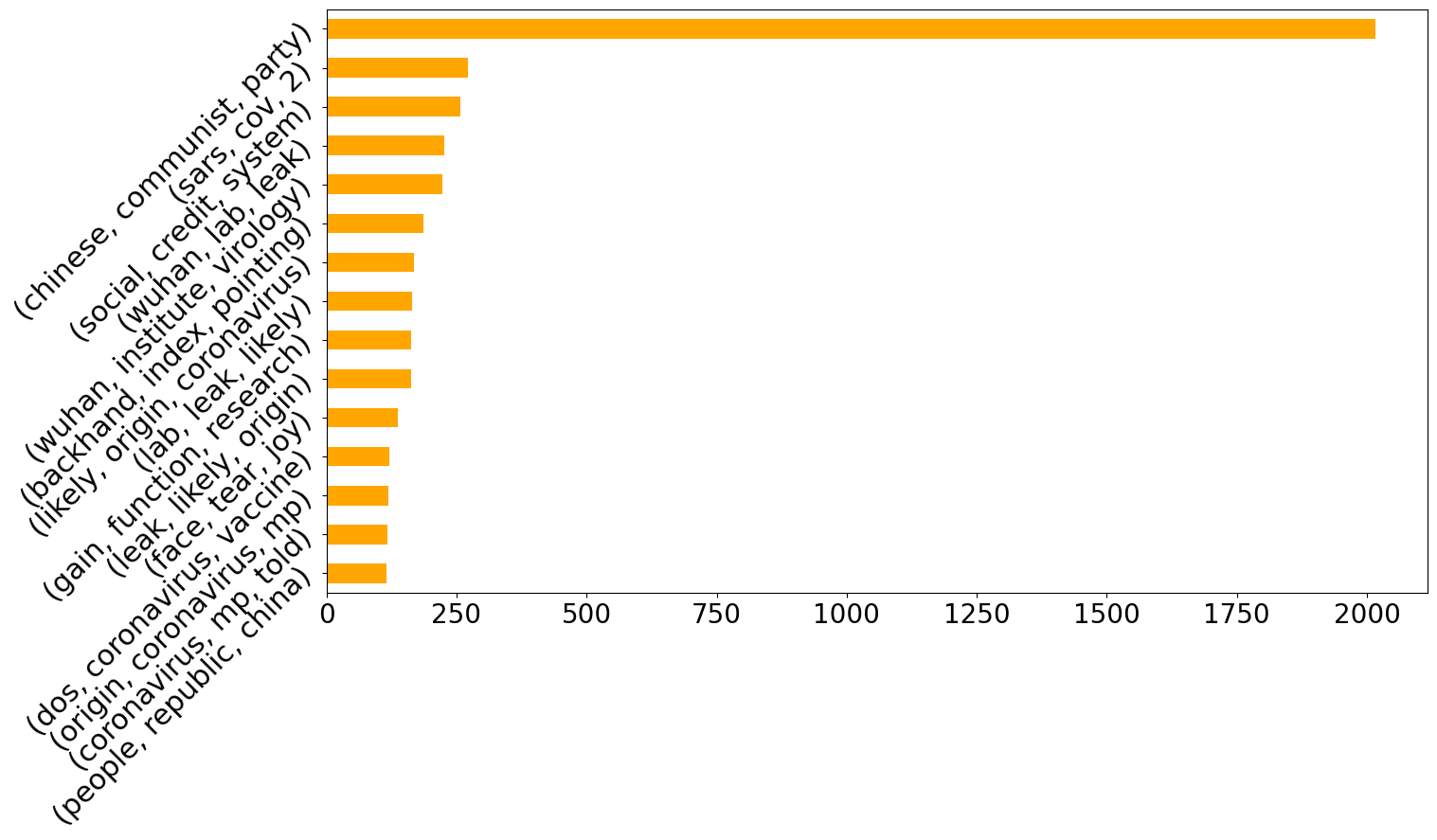}
\caption{United Kingdom} \label{fig:f}
\end{subfigure}

\caption{Top 15 trigrams of Sinophobic tweets for different countries.} \label{fig:trigram_thumbnails}
\end{figure*}

Figure \ref{fig:trigram_thumbnails} presents the top 15 trigrams for the different six countries where most frequent trigrams are "chinese communist party" and  "wuhan institute virology", aligning with the findings from the worldwide trigram plot in Figure \ref{fig:bigram_trigram}-Panel (b).

In the case of Australia, other notable trigrams include "communist party virus", "wuhan lab funded," and "know virus release". In the case of India, besides the common trigrams, phrases such as  "chinese coronavirus vaccine" and "coronavirus vaccine china" are frequent, reflecting a significant interest in China's role in vaccine development and distribution. In the case of Indonesia, we observe a similar with India, where the trigrams such as "communist party virus" and "chinese coronavirus vaccine" appear frequently. In the case of Japan, apart from the common trigrams, "sinopharm coronavirus vaccine," "evidence coronavirus lab," and "lab developed bioweapon" are prominent. In Brazil, we observe trigrams "china virus facemask" and "child china virus" suggest a focus on the impact of the virus on children and the use of facemasks, linking these topics explicitly to China. In the United Kingdom, trigrams such as "wuhan lab leak" and "likely origin coronavirus" point to concerns about the lab origin theory. 

Overall, although there are common themes across all countries, specific regional narratives and concerns are evident in the trigrams. Asian countries show a greater interest in Chinese vaccines, reflecting the importance of vaccination in their pandemic response. On the other hand, Australia and the United Kingdom have more Sinophobic-related trigrams among the top 15 compared to other countries, indicating a stronger focus on narratives that link the virus explicitly to China.

\subsection{Sentiment Analysis using BERT}

We evaluate the performance of the BERT model for sentiment analysis of Sinophobic tweets as shown in Table \ref{tab:metrics} which provides a detailed summary of performance metrics. 

\begin{table}[htbp!]
\centering
\begin{tabular}{lc}
\toprule
\textbf{Metric} & \textbf{Score} \\
\midrule
Hamming Loss & 0.1396 \\
Jaccard Score & 0.5138 \\
Label Ranking Average Precision Score & 0.7714 \\
F1 Macro Score & 0.5228 \\
F1 Micro Score & 0.5884 \\
\bottomrule
\end{tabular}
\caption{Performance metrics for the model.}
\label{tab:metrics}
\end{table}


The evaluation metrics indicate that our fine-tuned BERT model performs moderately well in identifying and ranking sentiments within the Sinophobic tweets. The low Hamming Loss and relatively high LRAP Score suggest that the model is effective at accurately predicting multiple relevant sentiments per tweet. The Jaccard, F1 Macro, and F1 Micro-Scores show that  the model successfully captures significant aspects of sentiment, making it a valuable tool for understanding the emotional undertones of Sinophobic discourse during the COVID-19 pandemic.

\subsection{Model Prediction}

We use the BERT-based refined sentiment analysis model for sentiment classification on the extracted Sinophobic tweets.

\begin{figure}[htbp!]
    \centering
    \includegraphics[width=0.49\textwidth]{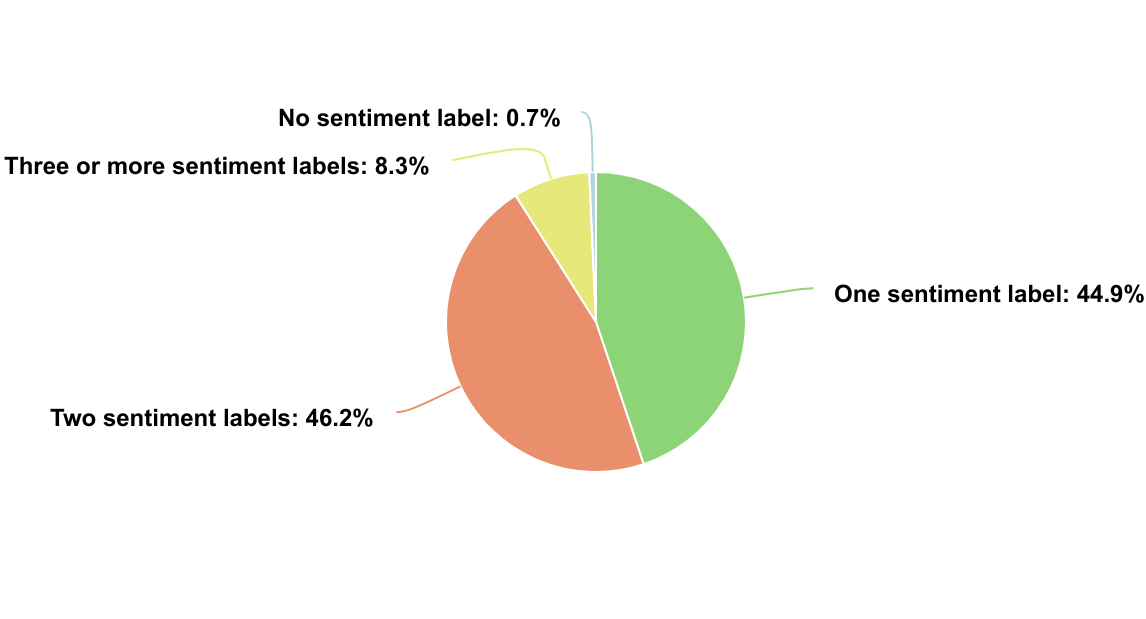}
    
    \caption{Percentage of Sinophobic tweets with the different number of labels.}
    \label{fig:pielabels}
\end{figure}

Figure \ref{fig:pielabels} illustrates the distribution of sentiment labels assigned to each tweet based on the number of sentiment labels for the entire dataset of Sinophobic tweets (including all the counted selected). The pie chart indicates that 0.7\% of the tweets have no sentiment label, 44.9\% have one sentiment label, 46.2\% have two sentiment labels, and 8.3\% have three or more sentiment labels.

\begin{figure}[htbp!]
    \centering
    \includegraphics[width=0.49\textwidth]{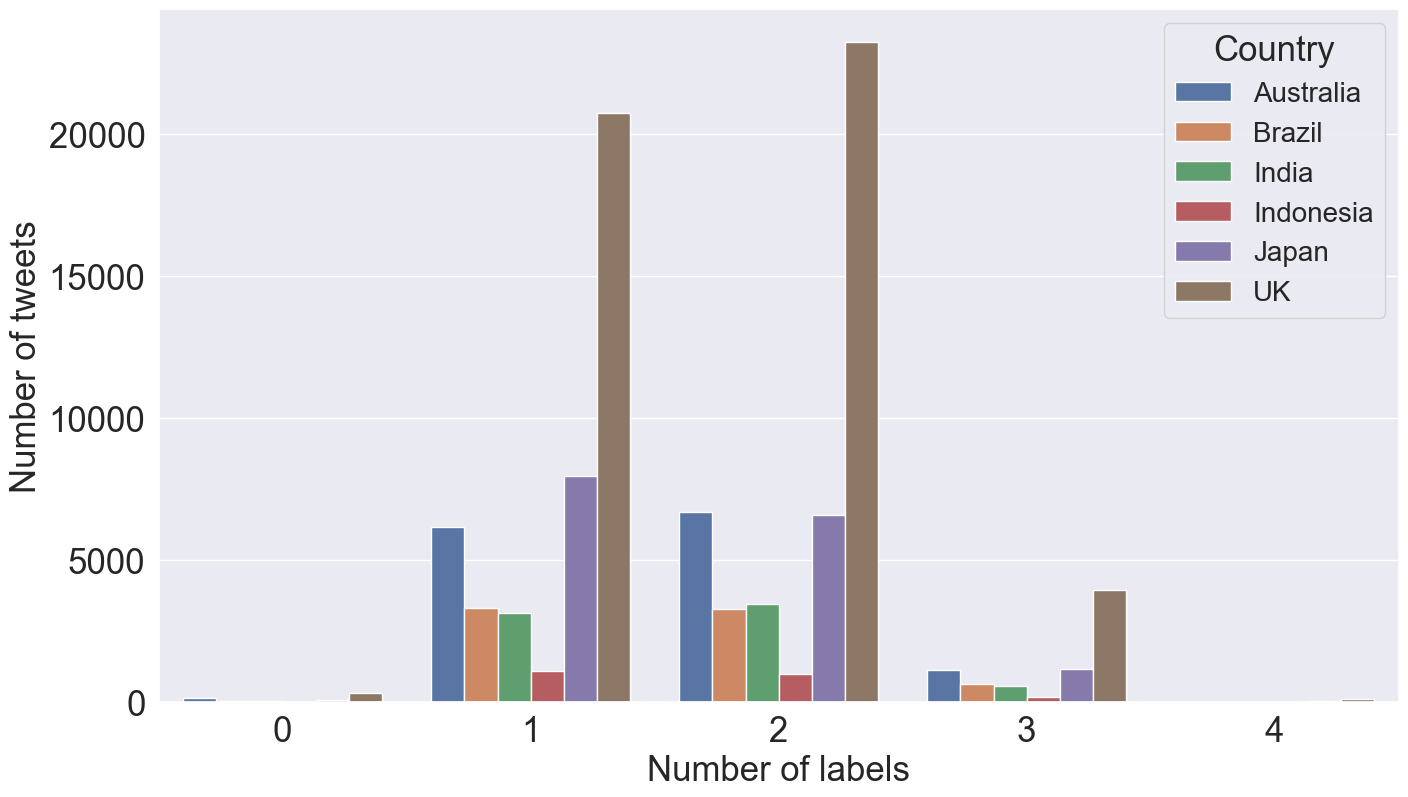}
    
    \caption{Number of Sinophobic tweets with different number of labels}
    \label{fig:barlables}
\end{figure}

Figure \ref{fig:barlables} provides a detailed view of the number of sentiment labels assigned to tweets across different countries. The bar graph reveals that the distribution patterns are quite similar across all the countries. The majority of tweets from each country are assigned one or two sentiment labels. The United Kingdom, Australia, and India have more tweets with two sentiment labels than one, indicating a more complex sentiment expression in these regions.

\subsection{Sentiment Analysis for different countries}

\begin{figure}[htbp!]
    \centering
    \includegraphics[width=\linewidth]{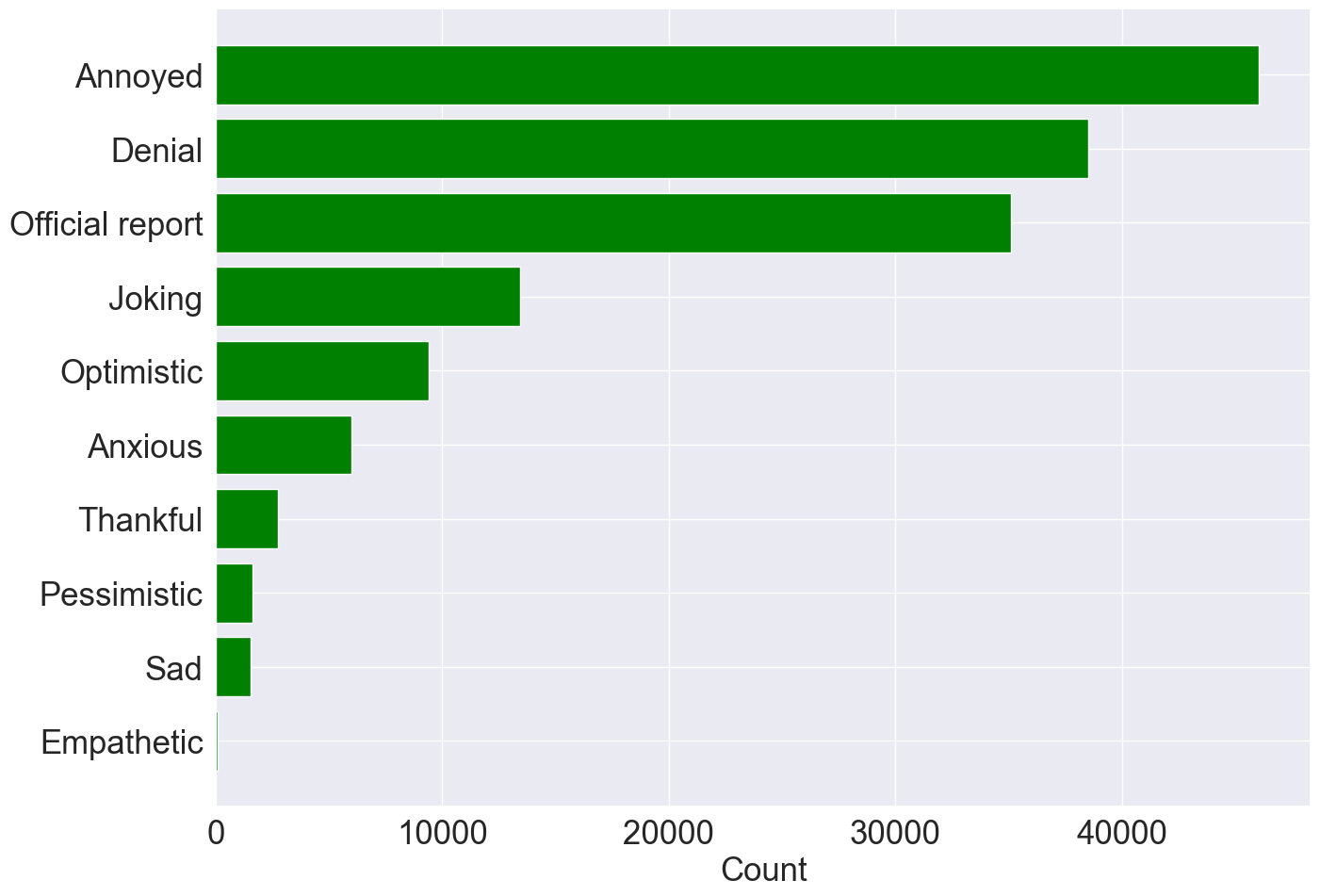}
    
    \caption{Total number of Sinophobic tweets for each sentiment.}
    \label{fig:tweets-sentiment}
\end{figure}

Next, we conduct a longitudinal analysis of the sentiments expressed in tweets across different countries from 2020-04 - 2022-01. Firstly, we examine the total number of tweets with each sentiment, as illustrated in Figure \ref{fig:tweets-sentiment}. The most common sentiment is "annoyed," with over 40,000 tweets classified under this label. This is followed by "denial," with a slightly lower count, and "official report," which also has a significant number of tweets. The sentiments "joking" and "optimistic" follow next, with counts slightly lower but still substantial. Besides these labels, the occurrence of the label "Empathetic" is extremely low, with the number of tweets carrying it being close to zero.

\begin{figure}[htbp]
    \centering
    \begin{subfigure}[b]{0.49\textwidth}
        \includegraphics[width=\textwidth]{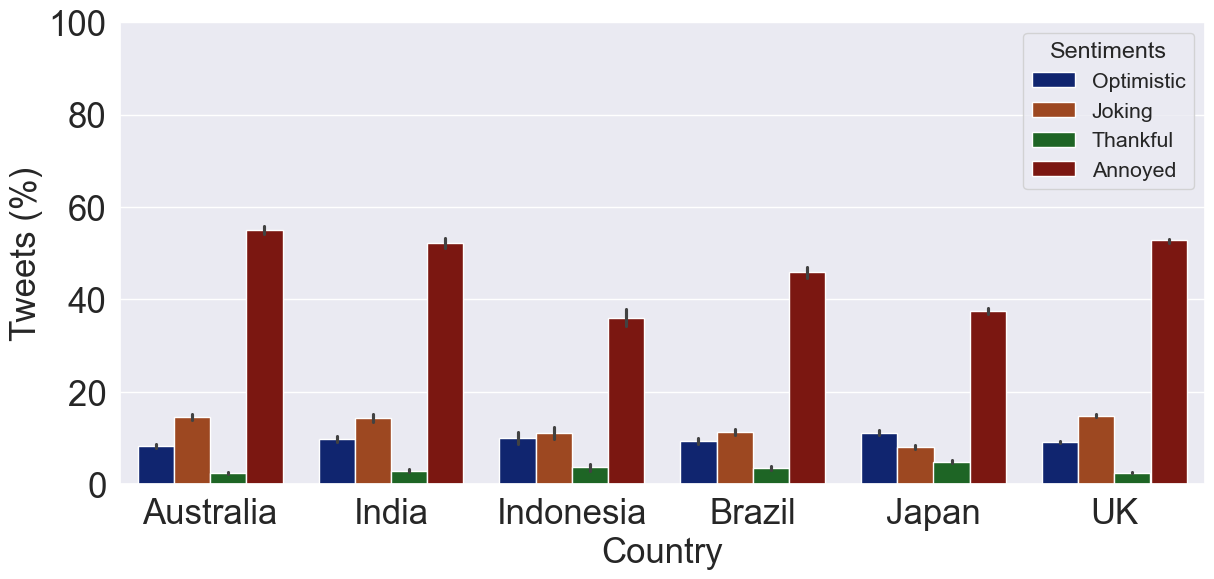}
        \caption{}
        \label{fig:bigram}
    \end{subfigure}
    \vfill
    \begin{subfigure}[b]{0.49\textwidth}
        \includegraphics[width=\textwidth]{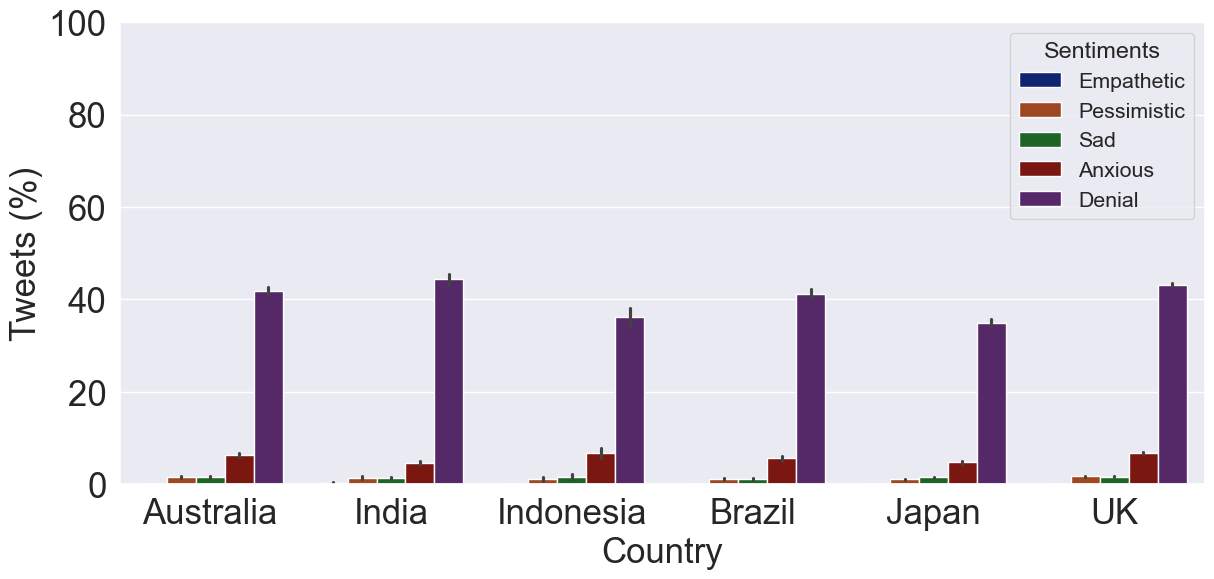}
        \caption{}
        \label{fig:trigram}
    \end{subfigure}
    \caption{Percentage of tweets of different sentiments for each country.}
    \label{fig:tweets-country}
\end{figure}

We excluded the "official report" for the rest of the analysis to focus on the emotional content of the tweets. Moving on to a country-wise analysis, Figure \ref{fig:tweets-country} depicts the percentage distribution of specific sentiments across different countries. In Figure \ref{fig:tweets-country}-Panel (a), the sentiments analysed include "optimistic", "joking", "thankful", and "annoyed". The data reveal that "annoyed" is the most dominant sentiment in all countries, with the highest percentage observed in Australia, India and the UK. The "Joking" sentiment also appears consistently across all countries but at a lower percentage compared to "annoyed". Figure \ref{fig:tweets-country}
-Panel (b) examines the distribution of "empathetic", "pessimistic", "sad", "anxious", and "denial" sentiments. We observe that "denial" emerges as the most frequent sentiment across all countries, especially in India, Australia and the United Kingdom. The "anxious" sentiment shows a notable presence, particularly in Indonesia and the United Kingdom. Despite the prevalence of negative sentiments, there are also notable occurrences of positive sentiments such as "optimistic" and "thankful". Among the countries analysed, we find that  Japan exhibits a higher proportion of tweets with positive sentiments.

\begin{figure}[htbp!]
    \centering
    \begin{subfigure}[b]{0.49\textwidth}
        \includegraphics[width=\textwidth]{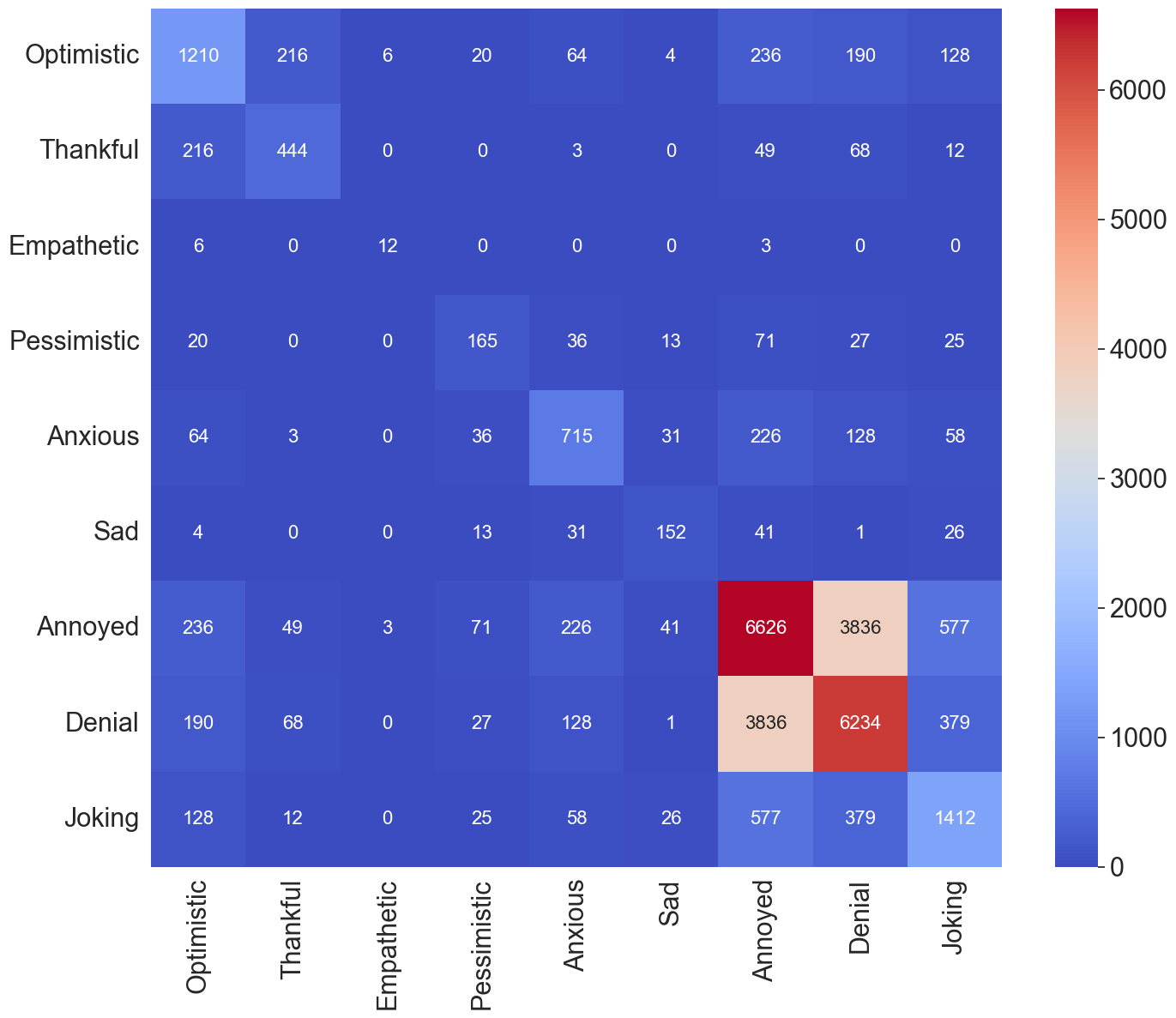}
        \caption{First year of COVD-19}
        \label{Bigram}
    \end{subfigure}
    \hfill
    \begin{subfigure}[b]{0.49\textwidth}
        \includegraphics[width=\textwidth]{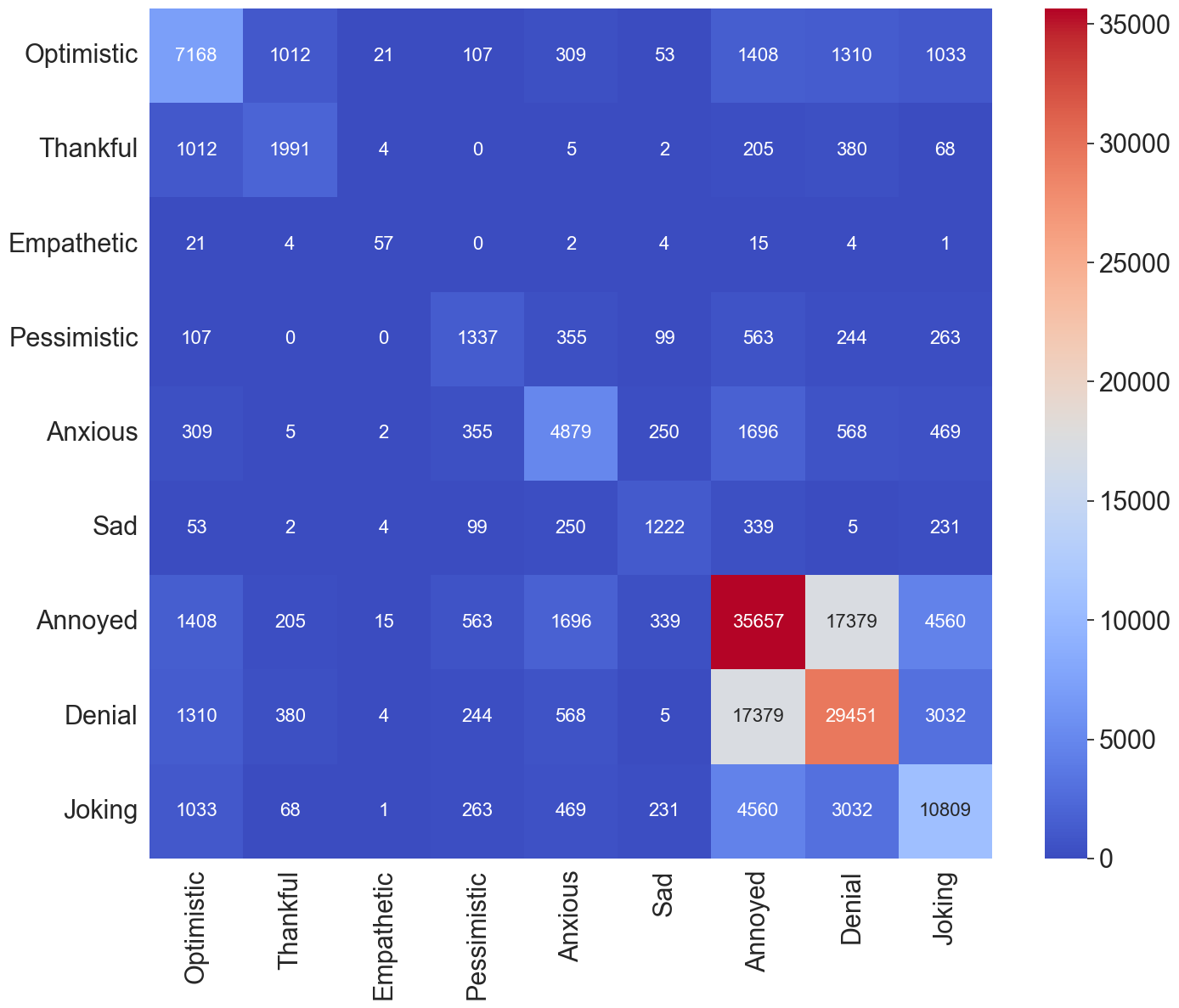}
        \caption{Second year of COVD-19}
        \label{Trigram}
    \end{subfigure}
    \caption{Heatmap of the first and second year of COVID-19 worldwide.}
    \label{fig:Heatmap}
\end{figure}

The two heatmaps in Figure \ref{fig:Heatmap} illustrate the frequency of co-occurring sentiments in the extracted tweets for the years 2020 and 2021, respectively. In 2020, as shown in Figure \ref{fig:Heatmap}-Panel (a), the most frequent individual sentiment is "annoyed", followed by "denial" and "joking". The highest co-occurrence is observed between "annoyed" and "denial," indicating that these sentiments are often expressed together. In 2021, as shown in Figure \ref{fig:Heatmap}-Panel (b), the trends are similar to 2020. However, there is an increase in the occurrence of these sentiments. The spikes in new COVID-19 cases in early 2021, particularly in India, as shown in Figure \ref{fig:COVIDcases}, correspond to an increase in the number of tweets expressing sentiments of "annoyed" and "denial", as observed in the heatmap for 2021.

\subsection{Polarity Score}

\begin{figure}[htbp!]
    \centering
    \begin{subfigure}[b]{0.49\textwidth}
        \includegraphics[width=\textwidth]{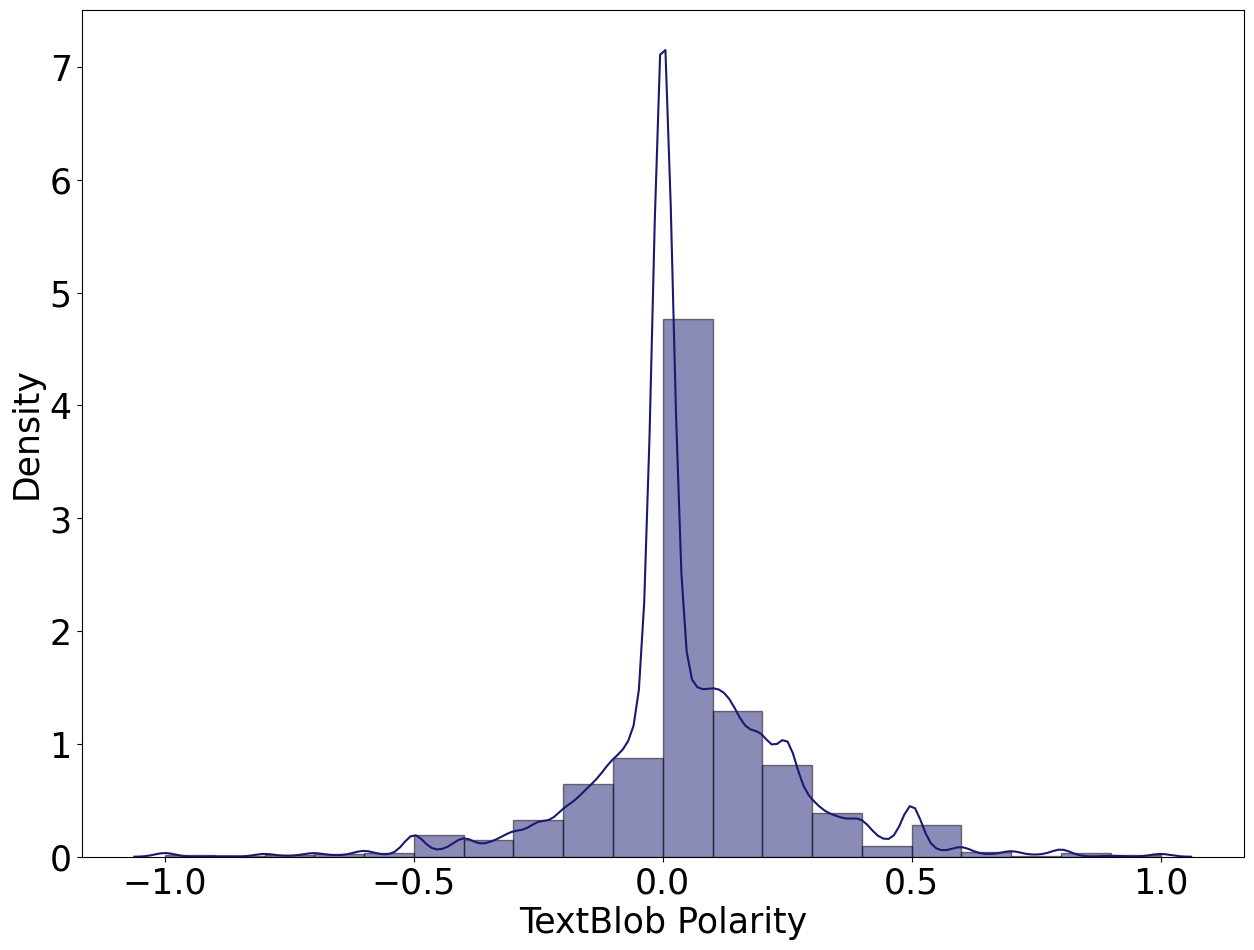}
        \caption{TextBlob}
        \label{Bigram}
    \end{subfigure}
    \hfill
    \begin{subfigure}[b]{0.49\textwidth}
        \includegraphics[width=\textwidth]{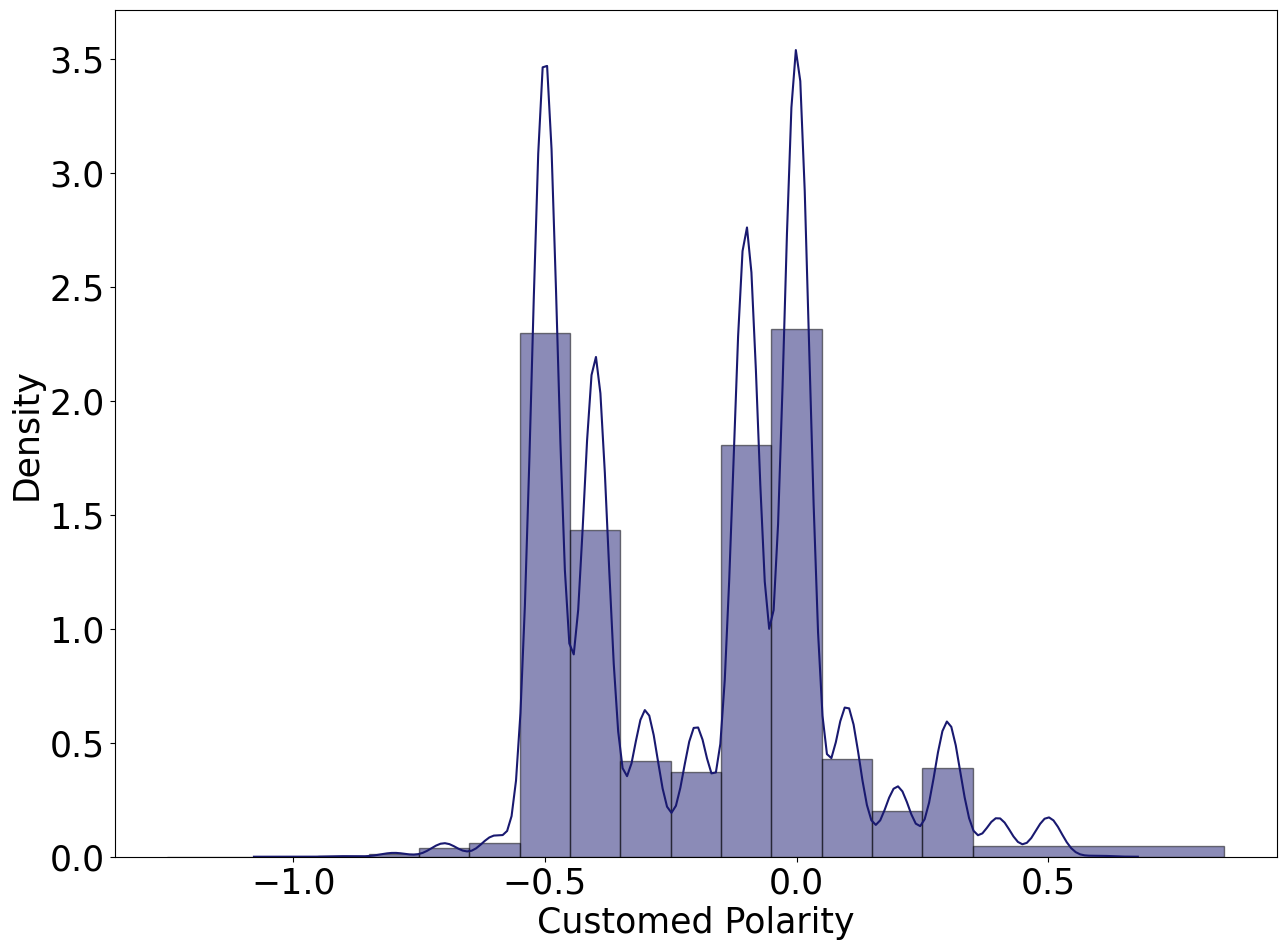}
        \caption{Custom weights}
        \label{Trigram}
    \end{subfigure}
    \caption{Distribution of TextBlob and Custom weight polarity scores.}
    \label{fig:dispolarity}
\end{figure}

We evaluate the polarity scores of the tweets using two different methods: TextBlob and a custom weight ratio approach. The polarity score is a measure of the sentiment expressed in a text, with positive scores indicating positive sentiment, negative scores indicating negative sentiment, and scores close to zero indicating neutral sentiment. TextBlob is a popular Python library for processing text using  tools for  NLP tasks. The custom weight ratio approach involves assigning weights to each sentiment, as shown in Table \ref{tab:weight}. This approach aims to capture more nuanced sentiment expressions that might be overlooked by generic sentiment analysis tools.
The distribution of TextBlob polarity scores for the dataset is shown in Figure \ref{fig:dispolarity}-Panel (a). From the figure, it is evident that the majority of the tweets have a polarity score close to zero, indicating neutral sentiment. The distribution of the custom weight ratio polarity scores is illustrated in Figure \ref{fig:dispolarity}-Panel (b). Similar to the TextBlob results, the custom method also shows a concentration of polarity scores around zero. However, the custom method reveals more granularity in the distribution, with several distinct peaks at various negative scores, especially -0.50. This indicates that the custom method might be capturing more subtle variations in sentiment.

\begin{figure}[htbp!]
    \centering
    \includegraphics[width=\linewidth]{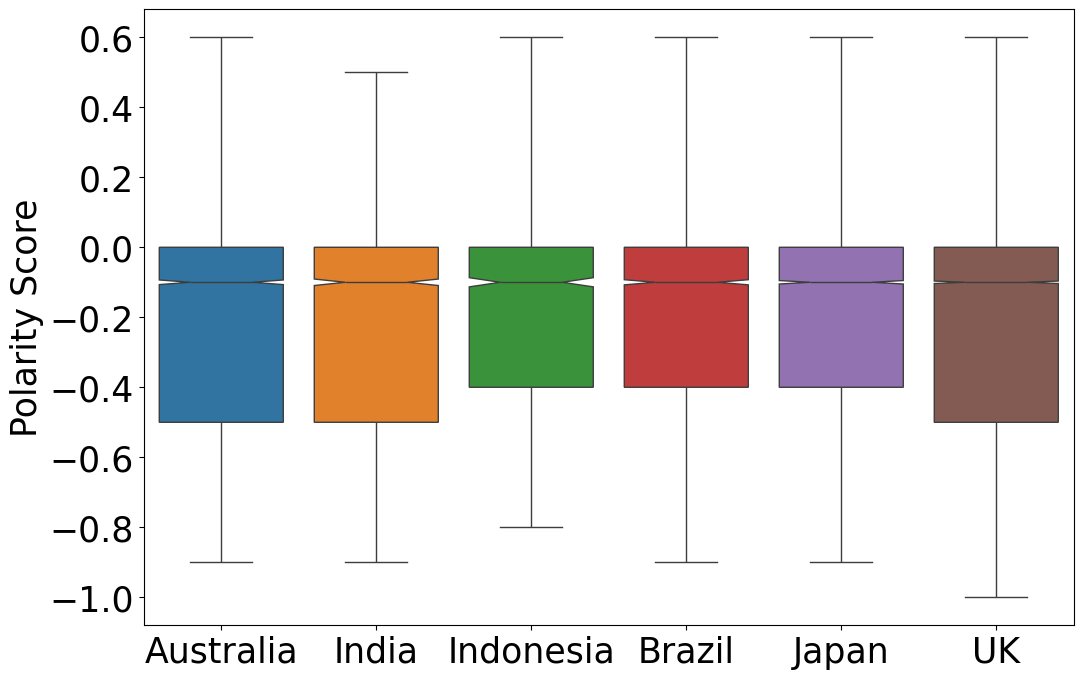}
    
    \caption{Box plot of custom weight polarity.}
    \label{fig:custompolarity}
\end{figure}

\begin{figure}[htbp!]
    \centering
    \includegraphics[width=\linewidth]{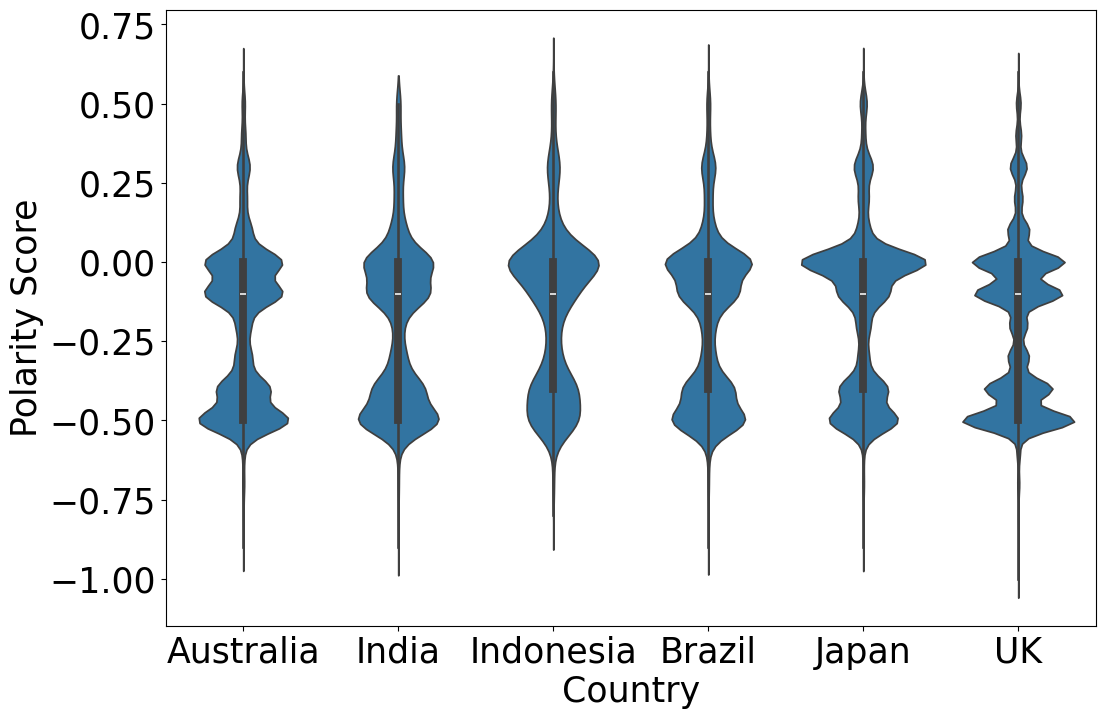}
    
    \caption{Violin plot of custom weight polarity.}
    \label{fig:violin}
\end{figure}

To further analyze the sentiment distribution across different countries, we plotted the polarity scores of tweets from six countries: Australia, India, Indonesia, Brazil, Japan, and the UK. Figure \ref{fig:custompolarity} illustrates the box plot of polarity scores for each country using the custom weight ratio method. The median polarity scores for all six countries are slightly below zero and the interquartile range boxes span from around -0.4 to 0.0, indicating a predominantly negative sentiment. Notably, Australia, India, and the UK have a lower bottom quartile (Q1) compared to other countries, suggesting a stronger negative sentiment in these regions. Figure \ref{fig:violin} presents the violin plot of polarity scores for the same six countries. The violin plots provide a detailed view of the distribution and density of the polarity scores.  The violin plot shows that most tweets have negative polarity scores which is consistent with the box plot. The density peaks at around -0.5 and 0.0 for all countries which further confirms the overall negative sentiment.

\begin{figure*}[htbp]
    \centering
    \small
    \begin{subfigure}[b]{0.9\textwidth}
        \includegraphics[width=\textwidth]{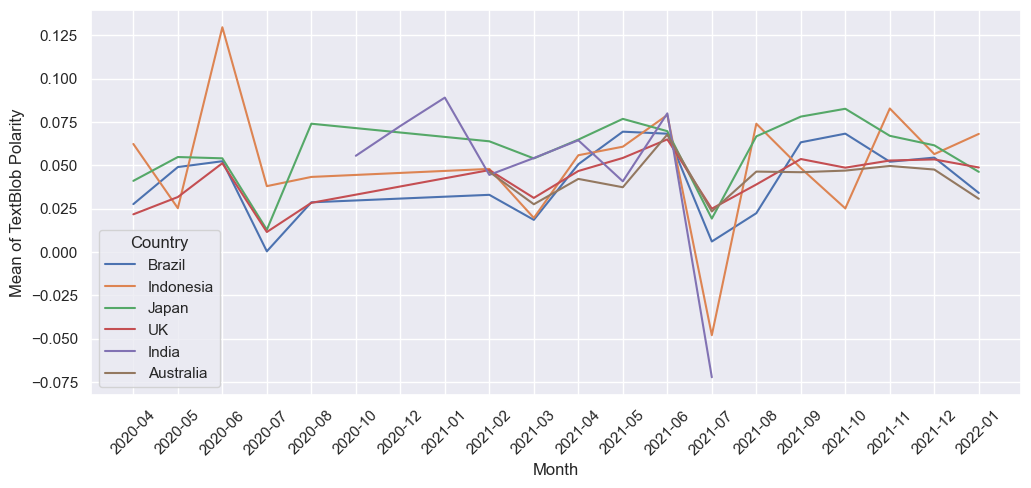}
        \caption{TextBlob polarity}
        \label{fig:bigram}
    \end{subfigure}
    \vfill
    \begin{subfigure}[b]{0.9\textwidth}
        \includegraphics[width=\textwidth]{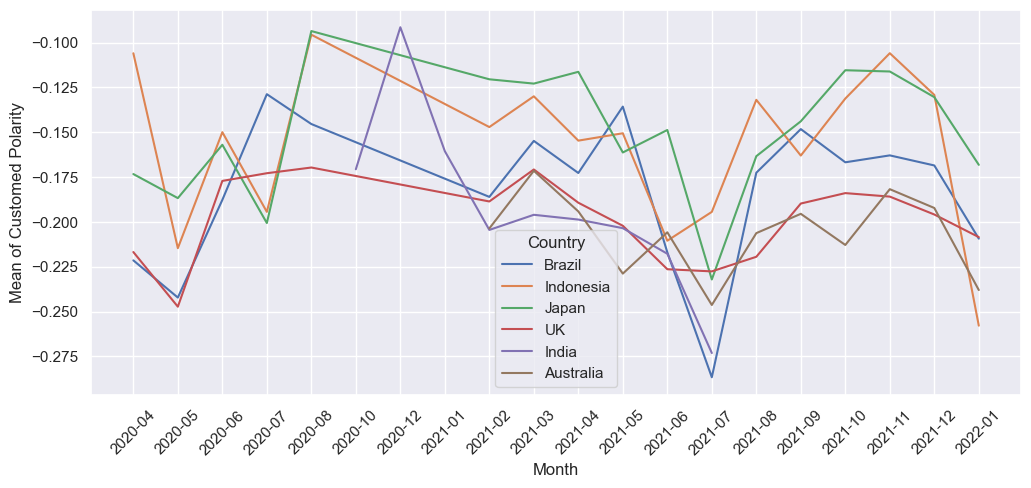}
        \caption{Custom weight polarity}
        \label{fig:trigram}
    \end{subfigure}
    \caption{Mean polarity of TextBlob and custom weight over month.}
    \label{fig:meanpolarity}
\end{figure*}

In Figure \ref{fig:meanpolarity}, we observe the mean polarity scores calculated using TextBlob and a custom-weighted approach from April 2020 to January 2022. The mean polarity score for each month and country is calculated by averaging the polarity scores of all tweets from that specific country within that month. Examining the TextBlob polarity scores (Figure \ref{fig:meanpolarity}-(a)), we observe an initial positive spike around May 2020, which coincides with the early stages of the pandemic when countries were implementing initial response measures. However, as the pandemic progressed, the sentiment became more negative. In comparison,  TextBlob generally yields higher (more positive) polarity scores than the custom-weighted approach (Figure \ref{fig:meanpolarity}-(b)), suggesting that the latter is more sensitive to negative sentiments. There is a noticeable correlation between the number of new COVID-19 cases and the polarity scores. Figure \ref{fig:COVIDcases} shows a significant peak in new cases around May 2021, followed by a steep decline and subsequent fluctuations. India experienced the highest spike in new COVID-19 cases during this period, which aligns with the observed dip in both methods  (Figure \ref{fig:meanpolarity}-(a) and (b)) around the same time. Conversely, countries with relatively stable COVID-19 case numbers, such as Japan and Australia, exhibit less fluctuation in their sentiment scores. This suggests that the evolution of the pandemic likely influenced public sentiment and the prevalence of Sinophobic tweets.

\begin{figure}[htbp!]
    \centering
    \begin{subfigure}[b]{0.49\textwidth}
        \includegraphics[width=\textwidth]{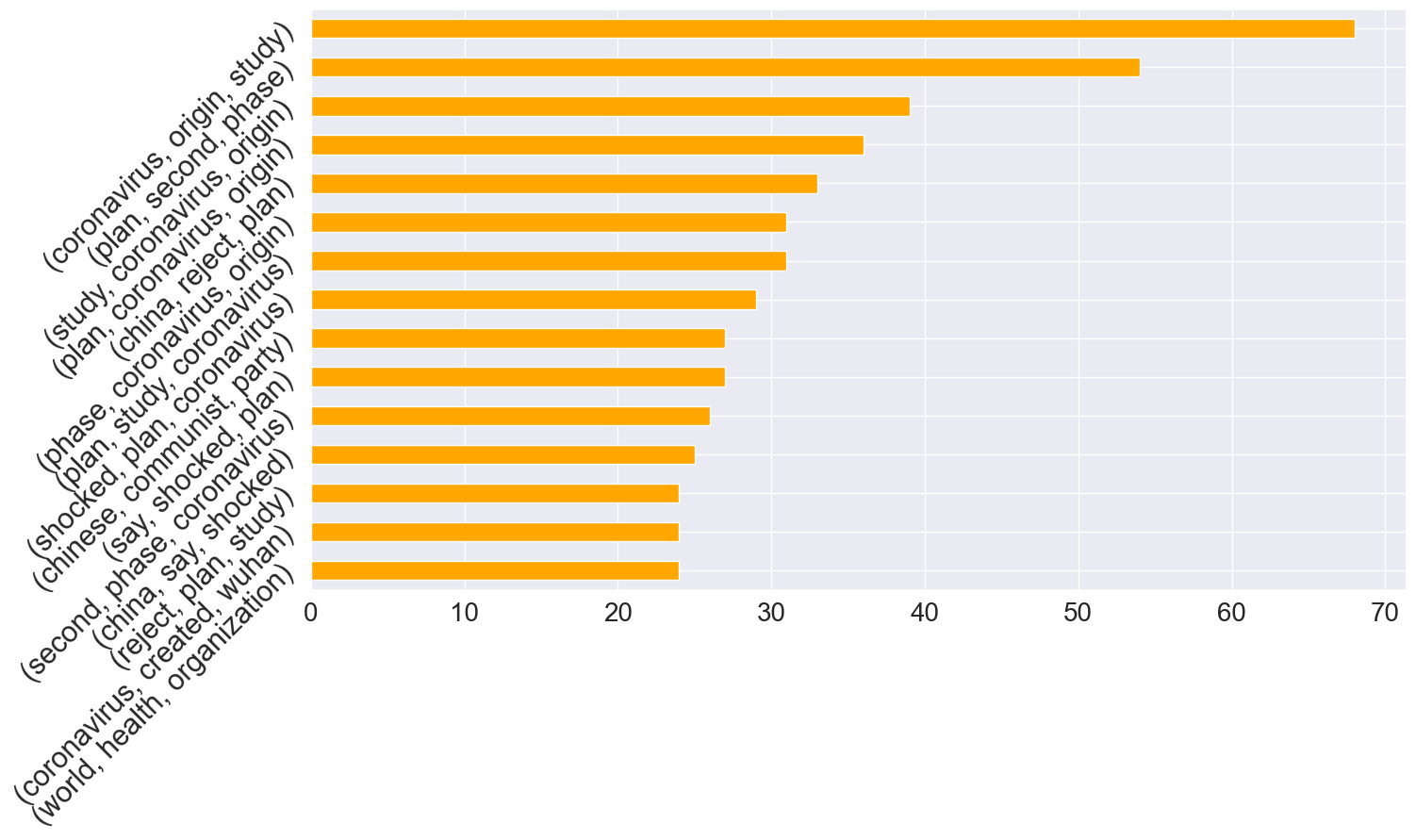}
        \caption{2021-07}
        \label{Bigram}
    \end{subfigure}
    \hfill
    \begin{subfigure}[b]{0.49\textwidth}
        \includegraphics[width=\textwidth]{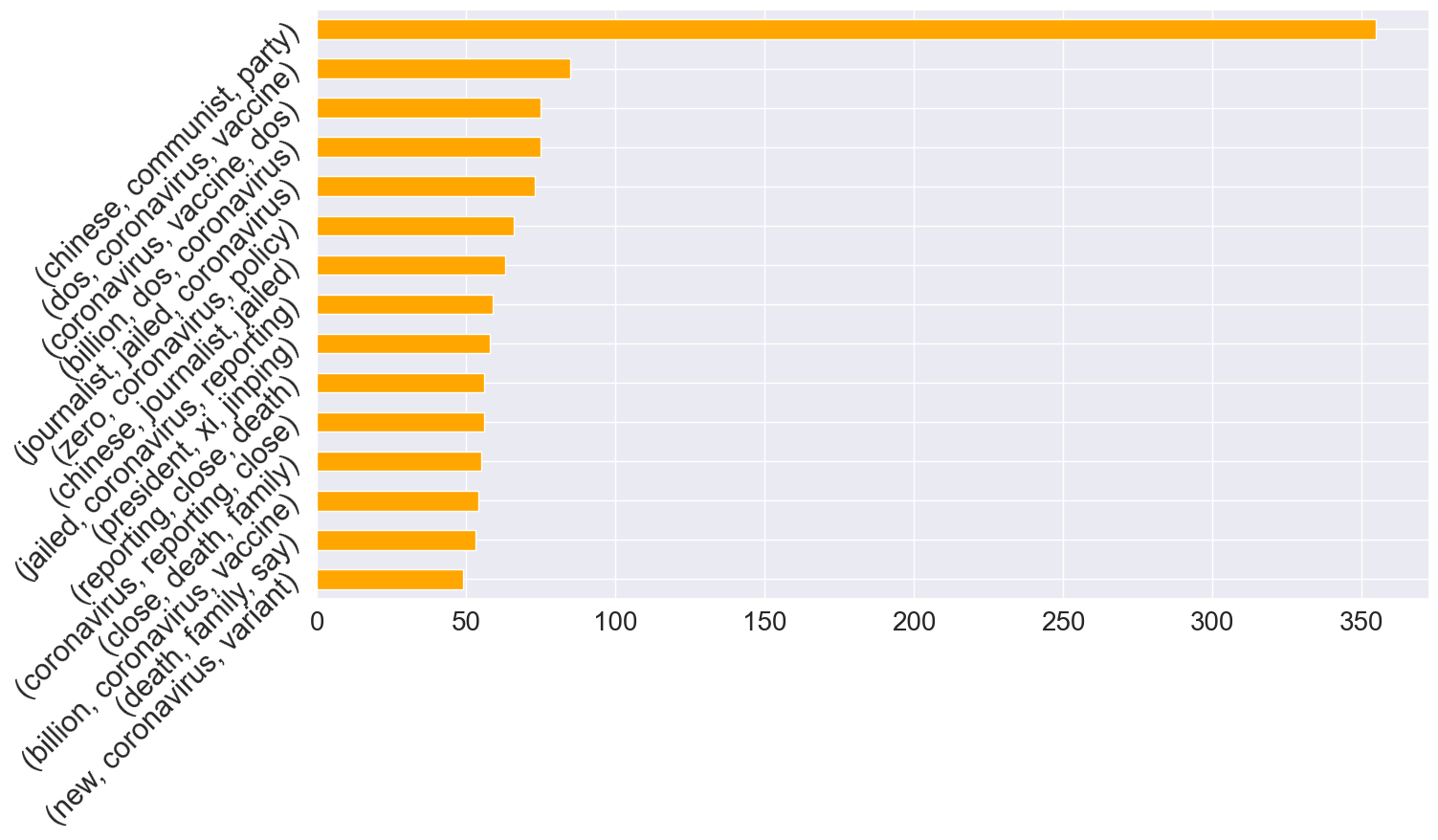}
        \caption{2021-11}
        \label{Trigram}
    \end{subfigure}
    \caption{Top 15 trigrams for 2021-07 and 2021-11 worldwide}
    \label{fig:TriTimezone}
\end{figure}

The analysis of the mean polarity over time in Figure \ref{fig:meanpolarity}-(b) reveals that July 2021 and November 2021 had notably low and high  polarity, respectively. 

Figure \ref{fig:TriTimezone} presents the top 15 trigrams for these months where we observe that in July 2021, the top trigrams include "coronavirus origin study", "plan second phase", "study coronavirus origin", and "china reject plan". These trigrams reflect the global focus on the investigation of the origins of COVID-19, particularly the second phase of studies and the controversies surrounding China's role and response. The prevalence of terms like "plan," "study," and "origin" suggests a period marked by heightened scrutiny and investigation into the virus's beginnings, contributing to the lower mean polarity observed. In contrast, November 2021 exhibits a higher polarity score. The top trigrams include "dos coronavirus vaccine", "coronavirus vaccine dos", and "billion dos coronavirus". This shift indicates a period focused more on vaccination efforts and the distribution of vaccines, with a significant emphasis on China's role in providing vaccines globally. The terms "journalist jailed coronavirus" and "chinese journalist jailed" aligns with news events, such as the imprisonment of a Chinese journalist for reporting on COVID-19 \cite{ohchr2021journalist}. Additionally, the terms "zero covid policy" and "new coronavirus variant" indicate ongoing public concerns about China's COVID-19 policies and the emergence of new variants.

Therefore, the rise in polarity score in November 2021 may be linked to the public attention shift from the origin of COVID-19 to vaccine distribution and global recovery efforts. As discussions increasingly shifted to vaccination progress and the management of the pandemic, the sentiment towards China may have been influenced by these broader global events, leading to a more positive outlook compared to earlier in the year.

Tables \ref{tab:positivesample} and  \ref{tab:negativesample} present several randomly selected sample tweets with positive or neutral and negative polarity scores, respectively.

\begin{table*}[htbp!]
\centering
\begin{tabular}{|>{\raggedright\arraybackslash}p{4cm}|>{\raggedright\arraybackslash}p{8cm}|>{\raggedright\arraybackslash}p{2cm}|>{\raggedright\arraybackslash}p{1.5cm}|}
\hline
\textbf{Sentiments} & \textbf{Sample tweet} & \textbf{Country} & \textbf{Score} \\
\hline
Optimistic, Joking & \textit{"I bought a subscription to the South China Morning Post yesterday. I was getting an excellent article or two from them each morning 
when checking out the Covid-19 situation.
\newline So, I thought enough of getting  articles  for free. Glad to learn that the SCMP is owned by Jack Ma."} & Australia & 0.4 \\
\hline
Joking & \textit{"Yay! A new virus??? Thanks China!\newline
Kung-Flu: The Legend Continues"} & Australia & 0.1 \\
\hline
Official report & \textit{China Emerges As Leader In Solar PV Installations As COVID-19 Makes Dent In Renewables Market} & India & 0.0 \\
\hline
Optimistic, Official report & \textit{"RT @Reuters: Crowds fill streets in China's pandemic-hit Wuhan, celebrate New Year"} & India & 0.3 \\
\hline
Joking & \textit{"RT @NikkeiAsia: THE VACCINE RACE $\vert$ WITH LOVE, FROM CHINA
\newline The Philippines begins COVID-19 vaccinations using shots donated by Beijing,"} & Indonesia & 0.1 \\
\hline
Optimistic, Annoyed & \textit{"RT @donof2050: @ChineseEmbinUK Dismantle all your artifical man made islands, then speak! \newline \#SouthChinaSea \#China @globaltimesnews"} & Indonesia & 0.2 \\ \hline
Optimistic, Thankful & \textit{"RT @business: China is doing a much better job sharing its Covid vaccine with the world than Western nations are"} & Brazil & 0.5 \\
\hline

Optimistic, Joking & \textit{"@MariaHernandor1 @sza\_jhcyto @mdlozanoe Unfortunately, also Brazilians, who were vaccinated with the Chinese covid vaccine, cannot enter in the vast majority of countries. Excellent meeting for you.  Opportunities will not be lacking."} & Brazil & 0.4 \\
\hline

Annoyed, Joking & \textit{"It's official. Ostraya's mining squillionaires are intergalactically high on CCP kool-aid."} & Japan & 0.0 \\
\hline

Thankful, Official report & \textit{"China's huge support to the multilateral environment agenda is key to hastening low carbon growth while shielding communities and ecosystems from shocks linked to the COVID-19 pandemic."} & Japan & 0.2 \\ \hline
Official report & \textit{"Major Chinese City Locks Down to Control Covid Outbreak"} & UK & 0.0 \\
\hline

Optimistic & \textit{"We’re back in China town again today, please call down if you want to get a vaccination \- the Mcr marathon is on today so plan your journey ahead \includegraphics[height=0.7em]{smile.png} Good luck to all the amazing runners"} & UK & 0.3 \\
\hline
\end{tabular}
\caption{Sentiment prediction outcomes with positive or neutral polarity scores for randomly selected tweet samples of different countries.}
\label{tab:positivesample}
\end{table*}

\begin{table*}[htbp!]
\centering
\small
\begin{tabular}{|>{\raggedright\arraybackslash}p{4cm}|>{\raggedright\arraybackslash}p{8cm}|>{\raggedright\arraybackslash}p{2cm}|>{\raggedright\arraybackslash}p{1.5cm}|}
\hline
\textbf{Sentiments} & \textbf{Sample tweet} & \textbf{Country} & \textbf{Score} \\
\hline
Annoyed, Denial & \textit{"China has no poverty, no sickness, no problems. Everyone is deliriously happy, the CCP is in complete control. They also tell a lot of lies."} & Australia & -0.5 \\
\hline
Denial & \textit{"@NSWHealth Doherty Institute receives \$US2 million donation from TikTok to fund COVID-19 research. My instinct tells me don’t believe the research. Oh did I mention CHINA?"} & Australia & -0.4 \\
\hline
Annoyed & \textit{"@SecBlinken 
 Actually Biden mental health is not good in Decision making.\newline
\#Cyber threat. China banned everything. Website Apps \newline Why we allow? CCP virus."} & India & -0.1 \\
\hline
Anxious & \textit{"RT @WIONews: \#Gravitas | Going by most estimates, a third wave of Wuhan virus cases could hit children."} & India & -0.2\\
\hline
Anxious, Annoyed & \textit{"RT @mako671178: [4 young men, demanding HK Govt to close the High-Speed Rail Station to prevent Chinese coronavirus]"} & Indonesia & -0.3 \\
\hline
Anxious, Annoyed, Denial & \textit{"RT @StephenMcDonell: Two years ago this happened in \#Wuhan. \#China \#COVID19"} & Indonesia & -0.7 \\
\hline
Anxious, Official report & \textit{"RT @EvanKirstel: \#China's Xi'an, which is currently on lockdown, launches city-wide disinfection as \#COVID cases continue to rise "} & Brazil & -0.2 \\
\hline

Annoyed, Denial, Joking & \textit{"Covid Cover-Up: New BOMBSHELL Lab Leak Report!! via @YouTube. I guess that I’m not the Conspiracy Tinfoil Hat Loon Theorist when in November 2019 that I said that it was a Wuhan Virology Lab Leak!
\newline It was because of a jacked HVAC System!!!"} & Brazil & -0.4 \\
\hline

Annoyed, Denail & \textit{"@globaltimesnews Anyway, China is the safest now, vaccinated for free. Moreover, Chinese vaccines can be vaccinated over 3 years old, unlike some provinces that claim to be countries to kill people."} & Japan & -0.5 \\
\hline

Annoyed & \textit{"\#USA won't repay debts to \#China? Or can't repay? \newline A good country, or a good person, is supposed to make every effort to repay debts. A cheeky, shameless \#debtor makes excuses to justify non-repayments"} & Japan & -0.1 \\
\hline

Pessimistic, Denial, Joking & \textit{"Surely China, as if covid19 was not enough, you now unleash a 22 ton rocket falling down in the next 2 and half hours? Location: unknown. Chinese rocket"} & UK & -0.6 \\
\hline

Denial & \textit{"@PeterDaszak This is such an interesting coincidence. And \#Fauci denies any \#GainOfFunction research with US funding. Interesting, as no WHO scientists could access all information from Wuhan lab. How does he know?"} & UK & -0.4 \\
 \hline
\end{tabular}
\caption{Sentiment prediction outcomes with negative polarity scores for randomly selected tweet samples of different countries.}
\label{tab:negativesample}
\end{table*}


\section{Discussion}

Our study on the sentiment analysis of Sinophobia during the COVID-19 pandemic has provided insightful observations into the dynamics of public sentiment across different countries. The findings reveal that the polarity scores of the sentiments are associated with the monthly new COVID-19 cases (Figure \ref{fig:COVIDcases}). As depicted in Figures \ref{fig:NumTweets} and  \ref{fig:meanpolarity}, the peaks in COVID-19 cases often coincided with spikes in the number of Sinophobic tweets and a drop in mean polarity scores, particularly during major outbreaks and public health interventions. In March 2021, The World Health Organization (WHO) released the results of the first investigation into the origins of COVID-19 \cite{world2021convened, smithsonian2021}. The speculation around the virus's origins, particularly the role of China, has significantly impacted public sentiment globally. This is evident from the significant drop after March 2021 in the polarity scores across various countries, as depicted in Figure \ref{fig:meanpolarity}-(b). Such events underscore the interplay between public health communications and social media sentiment, emphasising the need for transparent and accurate dissemination of information.


Our analysis of the predicted sentiment labels shows that a significant majority (91.1\%) of tweets were assigned one or two sentiment labels. This finding is consistent with the studies conducted by Chandra et al. \cite{chandra2021covid} on sentiment analysis during COVID-19 and  sentiment analysis of anti-vaccine tweets during COVID-19  \cite{chandra2023analysis}. This consistency suggests that the sentiment analysis models are robust across different datasets and contexts, capturing the predominant emotions expressed in social media discourse. 

Our sentiment analysis highlighted a predominant presence of negative sentiments towards China, with "annoyed" and "denial" being the most frequently expressed emotions, as shown in Figure \ref{fig:tweets-sentiment}. This trend is consistent across all selected countries, as illustrated in Figures \ref{fig:tweets-country}, where positive sentiments, though less frequent, were still present, with "optimistic" and "thankful" being notable examples.  One of the most surprising findings in these figures is the lack of "empathetic" sentiment which was present in previous studies, such as the case of COVID-19 sentiment analysis in India \cite{chandra2021covid}. Hence, empathy was expressed despite the catastrophic effect of COVID-19. However, it is predominantly absent when it comes to Sinophoic tweets, since tweets that contain hate speech and despise a community will lack empathy naturally.

An analysis of the bi-grams and trigrams in our dataset, shown in Figure \ref{fig:bigram_trigram}, reveals interesting trends in the language used in tweets related to Sinophobia. Notably, terms such as "Chinese communist" and "communist party" are among the most frequent, surpassing what might typically be expected phrases such as  "Wuhan virus" or "Chinese virus". This is surprising given that Viladrich \cite{viladrich2021sinophobic} highlighted how the widespread racial stigma was fuelled by terms such as the "Chinese virus". Additionally, our finding aligns with  Gao \cite{gao2022sinophobia} that Sinophobia is driven by not only the stereotypical association of Chinese people with the coronavirus but also by a complex phenomenon influenced by health, racial, and political factors.
The frequent use of politically charged terms reflects the discourse around COVID-19 and its origins was heavily politicised, and the significant impact of political narratives in shaping public sentiment during the pandemic.

Furthermore, the terms "Wuhan lab", "Wuhan Institute of Virology" and "Wuhan lab leak" appeared prominently among the bi-grams and trigrams, reflecting a specific focus on the origins of the virus. This aligns with the intense speculation and controversy surrounding the possibility of a lab leak as the source of COVID-19 \cite{maxmen2021covid}, which gained significant attention during the pandemic. The conspiracy theories surrounding these terms further fuelled public fear and suspicion, contributing to the negative sentiments observed.
 
 By comparing the Top 15 trigrams in July 2021 and November 2021, which are two months presenting notably low and high mean polarity scores, respectively. We noticed a significant psychological shift in people as the pandemic went on and appeared to be out of sight. People's focus transferred from "Why did this happen" to "when and how this can end". Moreover, from observing the high-frequency words (n-grams), it becomes evident how COVID-19 and Sinophobic sentiments might be politicised into tools for public opinion, deployed in power struggles between states.

The analysis of sample tweets, as shown in Table \ref{tab:positivesample} and Table \ref{tab:negativesample}, indicates that the model performs well in classifying sentiment accurately in most cases. The samples overall make sense and show the model's effectiveness in capturing the predominant sentiments expressed in the tweets. However, examining specific examples reveals some challenges in sentiment analysis, particularly those highlighted by Hussein \cite{hussein2018survey}, such as handling negation, spam and fake detection, and sarcasm detection. For instance, consider the second tweet in Table \ref{tab:positivesample}: \textit{"Yay! A new virus??? Thanks China! Kung-Flu: The Legend Continues"}. This tweet was classified as "Joking" and with a positive polarity score "0.1", but it is clearly an example of irony or sarcasm. In Table \ref{tab:negativesample}, \textit{"RT @EvanKirstel: \#China's Xi'an, which is currently on lockdown, launches city-wide disinfection as \#COVID cases continue to rise "} was labelled as "anxious" and "official report" with a polarity score "-0.2". Although the classification of "official report" is appropriate given the factual, news-like nature of the tweet, the inclusion of "anxious" reflects the underlying concern conveyed by the content. This aligns with the understanding that official statements during the pandemic, while intended to inform, often conveyed a sense of urgency and concern. These statements could be misinterpreted or manipulated to suit various narratives. Therefore, the need for transparent and cautious communication is critical to minimise misinformation and maintain public trust.

There are several limitations that need to be acknowledged. Firstly, both of our datasets are primarily sourced from X (Twitter), which may not fully represent the broad spectrum of public opinion. Twitter users tend to be younger and more technologically inclined \cite{sloan2015tweets}, which might not reflect the sentiments of the wider population. Additionally, the brevity of tweets (limited to 280 characters) can restrict the depth of expressed sentiments and the nuances of opinion, potentially leading to oversimplification in sentiment classification. As stated by Wankhade et al. \cite{wankhade2022survey}, the use of slang, abbreviations, and emojis in tweets may further affect the accuracy of sentiment analysis models. Secondly, as highlighted in the sample tweets, our model struggled with detecting sarcasm and irony, which are critical for accurate sentiment analysis. Moreover, the handling of negation and the detection of spam and fake content remain significant challenges that can impact the reliability of the sentiment classification. Thirdly, while we made efforts to filter and clean the dataset, the presence of noisy data, such as irrelevant tweets or misclassified sentiments, cannot be entirely ruled out. This noise can skew the analysis and affect the accuracy of the sentiment classification. For example, the tweet \textit{"Morrison is desperately trying to hide the covid flames that are well and truly out of control, whilst not admitting that they don't have enough PRC tests."} containing the phrase "prc test" was filtered as a Sinophobic tweet because "prc" stands for the People's Republic of China. However, in this context, it was likely a typo, and the user intended to refer to the "PCR test" for COVID-19.

In future research, we can expand the scope of data sources beyond Twitter to include other social media platforms like Facebook, Instagram, and Weibo could provide a more comprehensive understanding of public sentiment. Each platform has a different user demographic, and incorporating data from these sources would help capture a broader spectrum of opinions and emotions. Additionally, despite the robustness of BERT in sentiment analysis, future work could explore the use of more advanced large language models \cite{
thirunavukarasu2023large,chang2024survey}. By utilising other models, such as irony detection models, researchers can compare the performance and accuracy of different approaches in sentiment analysis. This comparison helps identify the strengths and weaknesses of each model, thereby informing the development of more sophisticated sentiment analysis tools. 
 Future work can also expand on the current study by longitudinal analysis of sentiments in entertainment media such as movies and music over the past few decades.  Finally, future sentiment analysis can be enhanced by incorporating cultural, religious, historical, political, and economic factors, which significantly influence public sentiment. Integrating these dimensions will provide a more nuanced understanding of the drivers behind Sinophobia during the COVID-19 pandemic.

\section{Conclusions}

The COVID-19 pandemic has significantly amplified xenophobia, particularly Sinophobia, resulting in widespread discrimination against individuals of Chinese descent. Our  study utilised deep learning-based language models to conduct sentiment analysis on social media data, focusing on Sinophobic sentiments expressed during the pandemic. By leveraging the SenWave dataset and fine-tuning a BERT model, we capture the nuances and evolution of these sentiments across the selected countries over time. The BERT-based model demonstrated moderate effectiveness, highlighting the need for advancements to better handle complex sentiments such as sarcasm and irony. The analysis revealed a correlation between spikes in Sinophobic tweets, drops in sentiment polarity scores and major COVID-19 outbreaks, indicating that the pandemic's progression fuelled Sinophobia. Negative sentiments like "Annoyed" and "Denial" were predominant, though some positive sentiments were also observed. Political narratives and misinformation, particularly linking COVID-19 to China, significantly impacted public sentiment.  Finally, there was a   lack of "empathetic" sentiment which was present in previous studies related to COVID-19 sentiment analysis; hence, empathy was expressed despite the catastrophic effect of COVID-19. Since  Sinophoic tweets contain hate speech and despise a community, the lack of empathy in such tweets is natural which highlights the way the political narratives in media viewed the pandemic and how it blamed the Chinese community.


\section*{Data and Code Avialbility} 

We provide open course code via Github repository \footnote{\url{https://github.com/sydney-machine-learning/sentimentanalysis-sinuphobiaCOVID19}}

 \bibliographystyle{elsarticle-num} 
 \bibliography{cas-refs}

\begin{thebibliography}{10}
\expandafter\ifx\csname url\endcsname\relax
  \def\url#1{\texttt{#1}}\fi
\expandafter\ifx\csname urlprefix\endcsname\relax\def\urlprefix{URL }\fi
\expandafter\ifx\csname href\endcsname\relax
  \def\href#1#2{#2} \def\path#1{#1}\fi

\bibitem{velavan2020covid}
T.~P. Velavan, C.~G. Meyer, The covid-19 epidemic, Tropical medicine \& international health 25~(3) (2020) 278.

\bibitem{yuki2020covid}
K.~Yuki, M.~Fujiogi, S.~Koutsogiannaki, Covid-19 pathophysiology: A review, Clinical immunology 215 (2020) 108427.

\bibitem{singhal2020review}
T.~Singhal, A review of coronavirus disease-2019 (covid-19), The Indian journal of pediatrics 87~(4) (2020) 281--286.

\bibitem{ciotti2020covid}
M.~Ciotti, M.~Ciccozzi, A.~Terrinoni, W.-C. Jiang, C.-B. Wang, S.~Bernardini, The covid-19 pandemic, Critical reviews in clinical laboratory sciences 57~(6) (2020) 365--388.

\bibitem{worldometer}
Worldometer, \href{https://www.worldometers.info/coronavirus/#countries}{Covid-19 coronavirus pandemic}, accessed: 2024-05-21 (2024).
\newline\urlprefix\url{https://www.worldometers.info/coronavirus/#countries}

\bibitem{clemente2021impact}
V.~J. Clemente-Su{\'a}rez, E.~Navarro-Jim{\'e}nez, L.~Moreno-Luna, M.~C. Saavedra-Serrano, M.~Jimenez, J.~A. Sim{\'o}n, J.~F. Tornero-Aguilera, The impact of the covid-19 pandemic on social, health, and economy, Sustainability 13~(11) (2021) 6314.

\bibitem{long2022covid}
E.~Long, S.~Patterson, K.~Maxwell, C.~Blake, R.~B. P{\'e}rez, R.~Lewis, M.~McCann, J.~Riddell, K.~Skivington, R.~Wilson-Lowe, et~al., Covid-19 pandemic and its impact on social relationships and health, J Epidemiol Community Health 76~(2) (2022) 128--132.

\bibitem{lopez2021more}
S.~Lopez-Leon, T.~Wegman-Ostrosky, C.~Perelman, R.~Sepulveda, P.~A. Rebolledo, A.~Cuapio, S.~Villapol, More than 50 long-term effects of covid-19: a systematic review and meta-analysis, Scientific reports 11~(1) (2021) 1--12.

\bibitem{gao2020mental}
J.~Gao, P.~Zheng, Y.~Jia, H.~Chen, Y.~Mao, S.~Chen, Y.~Wang, H.~Fu, J.~Dai, Mental health problems and social media exposure during covid-19 outbreak, Plos one 15~(4) (2020) e0231924.

\bibitem{tang2021important}
Q.~Tang, K.~Zhang, Y.~Li, The important role of social media during the covid-19 epidemic, Disaster medicine and public health preparedness 15~(4) (2021) e3--e4.

\bibitem{hussain2020role}
W.~Hussain, Role of social media in covid-19 pandemic, The International Journal of Frontier Sciences 4~(2) (2020) 59--60.

\bibitem{aggarwal2022role}
K.~Aggarwal, S.~K. Singh, M.~Chopra, S.~Kumar, Role of social media in the covid-19 pandemic: A literature review, Data mining approaches for big data and sentiment analysis in social media (2022) 91--115.

\bibitem{depoux2020pandemic}
A.~Depoux, S.~Martin, E.~Karafillakis, R.~Preet, A.~Wilder-Smith, H.~Larson, The pandemic of social media panic travels faster than the covid-19 outbreak (2020).

\bibitem{gao2022sinophobia}
Z.~Gao, Sinophobia during the covid-19 pandemic: Identity, belonging, and international politics, Integrative Psychological and Behavioral Science 56~(2) (2022) 472--490.

\bibitem{peyrouse2016discussing}
S.~Peyrouse, Discussing china: sinophilia and sinophobia in central asia, Journal of Eurasian Studies 7~(1) (2016) 14--23.

\bibitem{bille2017sinophobia}
F.~Bill{\'e}, Sinophobia: Anxiety, violence, and the making of Mongolian identity, University of Hawaii Press, 2017.

\bibitem{tahmasbi2021go}
F.~Tahmasbi, L.~Schild, C.~Ling, J.~Blackburn, G.~Stringhini, Y.~Zhang, S.~Zannettou, “go eat a bat, chang!”: On the emergence of sinophobic behavior on web communities in the face of covid-19, in: Proceedings of the web conference 2021, 2021, pp. 1122--1133.

\bibitem{viladrich2021sinophobic}
A.~Viladrich, Sinophobic stigma going viral: Addressing the social impact of covid-19 in a globalized world, American Journal of Public Health 111~(5) (2021) 876--880.

\bibitem{zhao2023survey}
W.~X. Zhao, K.~Zhou, J.~Li, T.~Tang, X.~Wang, Y.~Hou, Y.~Min, B.~Zhang, J.~Zhang, Z.~Dong, et~al., A survey of large language models, arXiv preprint arXiv:2303.18223 (2023).

\bibitem{chang2024survey}
Y.~Chang, X.~Wang, J.~Wang, Y.~Wu, L.~Yang, K.~Zhu, H.~Chen, X.~Yi, C.~Wang, Y.~Wang, et~al., A survey on evaluation of large language models, ACM Transactions on Intelligent Systems and Technology 15~(3) (2024) 1--45.

\bibitem{chowdhary2020natural}
K.~Chowdhary, K.~Chowdhary, Natural language processing, Fundamentals of artificial intelligence (2020) 603--649.

\bibitem{nadkarni2011natural}
P.~M. Nadkarni, L.~Ohno-Machado, W.~W. Chapman, Natural language processing: an introduction, Journal of the American Medical Informatics Association 18~(5) (2011) 544--551.

\bibitem{kasneci2023chatgpt}
E.~Kasneci, K.~Se{\ss}ler, S.~K{\"u}chemann, M.~Bannert, D.~Dementieva, F.~Fischer, U.~Gasser, G.~Groh, S.~G{\"u}nnemann, E.~H{\"u}llermeier, et~al., Chatgpt for good? on opportunities and challenges of large language models for education, Learning and individual differences 103 (2023) 102274.

\bibitem{chang2023survey}
Y.~Chang, X.~Wang, J.~Wang, Y.~Wu, L.~Yang, K.~Zhu, H.~Chen, X.~Yi, C.~Wang, Y.~Wang, et~al., A survey on evaluation of large language models, ACM Transactions on Intelligent Systems and Technology (2023).

\bibitem{zhang2023video}
H.~Zhang, X.~Li, L.~Bing, Video-llama: An instruction-tuned audio-visual language model for video understanding, arXiv preprint arXiv:2306.02858 (2023).

\bibitem{thirunavukarasu2023large}
A.~J. Thirunavukarasu, D.~S.~J. Ting, K.~Elangovan, L.~Gutierrez, T.~F. Tan, D.~S.~W. Ting, Large language models in medicine, Nature medicine 29~(8) (2023) 1930--1940.

\bibitem{kitamura2023chatgpt}
F.~C. Kitamura, Chatgpt is shaping the future of medical writing but still requires human judgment (2023).

\bibitem{floridi2020gpt}
L.~Floridi, M.~Chiriatti, Gpt-3: Its nature, scope, limits, and consequences, Minds and Machines 30 (2020) 681--694.

\bibitem{achiam2023gpt}
J.~Achiam, S.~Adler, S.~Agarwal, L.~Ahmad, I.~Akkaya, F.~L. Aleman, D.~Almeida, J.~Altenschmidt, S.~Altman, S.~Anadkat, et~al., Gpt-4 technical report, arXiv preprint arXiv:2303.08774 (2023).

\bibitem{devlin2018bert}
J.~Devlin, M.-W. Chang, K.~Lee, K.~Toutanova, Bert: Pre-training of deep bidirectional transformers for language understanding, arXiv preprint arXiv:1810.04805 (2018).

\bibitem{team2023gemini}
G.~Team, R.~Anil, S.~Borgeaud, Y.~Wu, J.-B. Alayrac, J.~Yu, R.~Soricut, J.~Schalkwyk, A.~M. Dai, A.~Hauth, et~al., Gemini: a family of highly capable multimodal models, arXiv preprint arXiv:2312.11805 (2023).

\bibitem{medhat2014sentiment}
W.~Medhat, A.~Hassan, H.~Korashy, Sentiment analysis algorithms and applications: A survey, Ain Shams engineering journal 5~(4) (2014) 1093--1113.

\bibitem{prabowo2009sentiment}
R.~Prabowo, M.~Thelwall, Sentiment analysis: A combined approach, Journal of Informetrics 3~(2) (2009) 143--157.

\bibitem{yang2020senwave}
Q.~Yang, H.~Alamro, S.~Albaradei, A.~Salhi, X.~Lv, C.~Ma, M.~Alshehri, I.~Jaber, F.~Tifratene, W.~Wang, et~al., Senwave: Monitoring the global sentiments under the covid-19 pandemic, arXiv preprint arXiv:2006.10842 (2020).

\bibitem{wankhade2022survey}
M.~Wankhade, A.~C.~S. Rao, C.~Kulkarni, A survey on sentiment analysis methods, applications, and challenges, Artificial Intelligence Review 55~(7) (2022) 5731--5780.

\bibitem{chandra2021covid}
R.~Chandra, A.~Krishna, Covid-19 sentiment analysis via deep learning during the rise of novel cases, PloS one 16~(8) (2021) e0255615.

\bibitem{lande2023deep}
J.~Lande, A.~Pillay, R.~Chandra, Deep learning for covid-19 topic modelling via twitter: Alpha, delta and omicron, PloS one 18~(8) (2023) e0288681.

\bibitem{chandra2023analysis}
R.~Chandra, J.~Sonawane, J.~Lande, C.~Yu, An analysis of vaccine-related sentiments from development to deployment of covid-19 vaccines, arXiv preprint arXiv:2306.13797 (2023).

\bibitem{krumpal2013determinants}
I.~Krumpal, Determinants of social desirability bias in sensitive surveys: a literature review, Quality \& quantity 47~(4) (2013) 2025--2047.

\bibitem{samuelson2023generative}
P.~Samuelson, Generative ai meets copyright, Science 381~(6654) (2023) 158--161.

\bibitem{kaddour2023challenges}
J.~Kaddour, J.~Harris, M.~Mozes, H.~Bradley, R.~Raileanu, R.~McHardy, Challenges and applications of large language models, arXiv preprint arXiv:2307.10169 (2023).

\bibitem{hadi2023large}
M.~U. Hadi, R.~Qureshi, A.~Shah, M.~Irfan, A.~Zafar, M.~B. Shaikh, N.~Akhtar, J.~Wu, S.~Mirjalili, et~al., Large language models: a comprehensive survey of its applications, challenges, limitations, and future prospects, Authorea Preprints (2023).

\bibitem{hussein2018survey}
D.~M. E.-D.~M. Hussein, A survey on sentiment analysis challenges, Journal of King Saud University-Engineering Sciences 30~(4) (2018) 330--338.

\bibitem{zhan2024optimization}
T.~Zhan, C.~Shi, Y.~Shi, H.~Li, Y.~Lin, Optimization techniques for sentiment analysis based on llm (gpt-3), arXiv preprint arXiv:2405.09770 (2024).

\bibitem{deng2023llms}
X.~Deng, V.~Bashlovkina, F.~Han, S.~Baumgartner, M.~Bendersky, Llms to the moon? reddit market sentiment analysis with large language models, in: Companion Proceedings of the ACM Web Conference 2023, 2023, pp. 1014--1019.

\bibitem{chen2020anti}
H.~A. Chen, J.~Trinh, G.~P. Yang, Anti-asian sentiment in the united states--covid-19 and history, The American Journal of Surgery 220~(3) (2020) 556--557.

\bibitem{masters2020covid}
T.~C. Masters-Waage, N.~Jha, J.~Reb, Covid-19, coronavirus, wuhan virus, or china virus? understanding how to “do no harm” when naming an infectious disease, Frontiers in psychology 11 (2020) 561270.

\bibitem{cheah2020covid}
C.~S. Cheah, C.~Wang, H.~Ren, X.~Zong, H.~S. Cho, X.~Xue, Covid-19 racism and mental health in chinese american families, Pediatrics 146~(5) (2020).

\bibitem{diggle2002analysis}
P.~Diggle, Analysis of longitudinal data, Oxford university press, 2002.

\bibitem{liu2016longitudinal}
Y.~Liu, S.~Mo, Y.~Song, M.~Wang, Longitudinal analysis in occupational health psychology: A review and tutorial of three longitudinal modeling techniques, Applied Psychology 65~(2) (2016) 379--411.

\bibitem{neufeld1981clinical}
V.~Neufeld, G.~Norman, J.~Feightner, H.~Barrows, Clinical problem-solving by medical students: a cross-sectional and longitudinal analysis, Medical education 15~(5) (1981) 315--322.

\bibitem{newsom2013longitudinal}
J.~Newsom, R.~N. Jones, S.~M. Hofer, Longitudinal data analysis: A practical guide for researchers in aging, health, and social sciences, Routledge, 2013.

\bibitem{lucas2020longitudinal}
C.~Lucas, P.~Wong, J.~Klein, T.~B. Castro, J.~Silva, M.~Sundaram, M.~K. Ellingson, T.~Mao, J.~E. Oh, B.~Israelow, et~al., Longitudinal analyses reveal immunological misfiring in severe covid-19, Nature 584~(7821) (2020) 463--469.

\bibitem{wang2020longitudinal}
C.~Wang, R.~Pan, X.~Wan, Y.~Tan, L.~Xu, R.~S. McIntyre, F.~N. Choo, B.~Tran, R.~Ho, V.~K. Sharma, et~al., A longitudinal study on the mental health of general population during the covid-19 epidemic in china, Brain, behavior, and immunity 87 (2020) 40--48.

\bibitem{chandra2024global}
R.~Chandra, J.~Lande, \href{https://dx.doi.org/10.2139/ssrn.4895603}{Global covid-19 x(twitter) dataset}, SSRNAccessed: 01-07-2024 (2024).
\newline\urlprefix\url{https://dx.doi.org/10.2139/ssrn.4895603}

\bibitem{senwave2024}
G.~Qiang, \href{https://github.com/gitdevqiang/SenWave}{Senwave: A large-scale dataset for social media analysis of covid-19}, accessed: 2024-06-24 (2024).
\newline\urlprefix\url{https://github.com/gitdevqiang/SenWave}

\bibitem{chandra2021biden}
R.~Chandra, R.~Saini, Biden vs trump: modeling us general elections using bert language model, IEEE access 9 (2021) 128494--128505.

\bibitem{koroteev2021bert}
M.~Koroteev, Bert: a review of applications in natural language processing and understanding, arXiv preprint arXiv:2103.11943 (2021).

\bibitem{miranda2023exploring}
C.~H. Miranda, G.~Sanchez-Torres, D.~Salcedo, Exploring the evolution of sentiment in spanish pandemic tweets: A data analysis based on a fine-tuned bert architecture, Data 8~(6) (2023) 96.

\bibitem{chalkidis2020legal}
I.~Chalkidis, M.~Fergadiotis, P.~Malakasiotis, N.~Aletras, I.~Androutsopoulos, Legal-bert: The muppets straight out of law school, arXiv preprint arXiv:2010.02559 (2020).

\bibitem{shao2020bert}
Y.~Shao, J.~Mao, Y.~Liu, W.~Ma, K.~Satoh, M.~Zhang, S.~Ma, Bert-pli: Modeling paragraph-level interactions for legal case retrieval., in: IJCAI, 2020, pp. 3501--3507.

\bibitem{rasmy2021med}
L.~Rasmy, Y.~Xiang, Z.~Xie, C.~Tao, D.~Zhi, Med-bert: pretrained contextualized embeddings on large-scale structured electronic health records for disease prediction, NPJ digital medicine 4~(1) (2021) 86.

\bibitem{liu2021med}
N.~Liu, Q.~Hu, H.~Xu, X.~Xu, M.~Chen, Med-bert: A pretraining framework for medical records named entity recognition, IEEE Transactions on Industrial Informatics 18~(8) (2021) 5600--5608.

\bibitem{kingma2014adam}
D.~P. Kingma, J.~Ba, Adam: A method for stochastic optimization, arXiv preprint arXiv:1412.6980 (2014).

\bibitem{chandra2022semantic}
R.~Chandra, V.~Kulkarni, Semantic and sentiment analysis of selected bhagavad gita translations using bert-based language framework, IEEE Access 10 (2022) 21291--21315.

\bibitem{dong2020interactive}
E.~Dong, H.~Du, L.~Gardner, An interactive web-based dashboard to track covid-19 in real time, The Lancet infectious diseases 20~(5) (2020) 533--534.

\bibitem{brown1992class}
P.~F. Brown, V.~J. Della~Pietra, P.~V. Desouza, J.~C. Lai, R.~L. Mercer, Class-based n-gram models of natural language, Computational linguistics 18~(4) (1992) 467--480.

\bibitem{bbc2021labLeakTheory}
B.~News, \href{https://www.bbc.com/news/world-asia-china-57268111}{Covid: Wuhan lab leak theory is 'feasible', says who team leader}, accessed: 2024-07-27 (2021).
\newline\urlprefix\url{https://www.bbc.com/news/world-asia-china-57268111}

\bibitem{ohchr2021journalist}
{OHCHR}, \href{https://www.ohchr.org/en/press-releases/2021/11/china-journalist-jailed-covid-reporting-seriously-ill-must-be-released-un}{China: Journalist jailed for covid reporting seriously ill, must be released - un} (2021).
\newline\urlprefix\url{https://www.ohchr.org/en/press-releases/2021/11/china-journalist-jailed-covid-reporting-seriously-ill-must-be-released-un}

\bibitem{world2021convened}
W.~H. Organization, et~al., Who-convened global study of origins of sars-cov-2: China part (2021).

\bibitem{smithsonian2021}
T.~Machemer, \href{https://www.smithsonianmag.com/smart-news/key-takeaways-who-report-origin-covid-19-180977423/}{Key takeaways from the who report on the origin of covid-19}, accessed: 2024-07-23 (2021).
\newline\urlprefix\url{https://www.smithsonianmag.com/smart-news/key-takeaways-who-report-origin-covid-19-180977423/}

\bibitem{maxmen2021covid}
A.~Maxmen, S.~Mallapaty, The covid lab-leak hypothesis: what scientists do and don’t know, Nature 594~(7863) (2021) 313--315.

\bibitem{sloan2015tweets}
L.~Sloan, J.~Morgan, P.~Burnap, M.~Williams, Who tweets? deriving the demographic characteristics of age, occupation and social class from twitter user meta-data, PloS one 10~(3) (2015) e0115545.

\end{thebibliography}





\end{document}